\documentclass[11pt]{article}
\usepackage[]{acl}
\usepackage{times}
\usepackage{latexsym}
\usepackage[T1]{fontenc}
\usepackage[utf8]{inputenc}
\usepackage{microtype}
\usepackage{inconsolata}
\usepackage{graphicx}
\usepackage{pgfplots}
\pgfplotsset{compat=1.18}
\usetikzlibrary{patterns}
\usepackage{booktabs}
\usepackage{multirow}
\usepackage{array}
\usepackage[table]{xcolor}
\usepackage{tabularx}
\usepackage{adjustbox}
\usepackage{amsmath}
\usepackage{amssymb}
\usepackage{xurl}
\usepackage{fvextra}
\usepackage{float}
\usepackage{placeins}
\usepackage{capt-of}
\usepackage[most]{tcolorbox}
\newcommand{\tablegroupdash}{\tikz[baseline]{\draw[line width=0.25pt,dash pattern=on 1.5pt off 1.5pt] (0,-\dp\strutbox)--(0,\ht\strutbox);}}
\newcommand{\method}{\textsc{Moira}}
\newcommand{\ls}{\ifmmode\text{\scshape Long--Short}\else\textsc{Long--Short}\fi}
\renewcommand{\sl}{\ifmmode\text{\scshape Short--Long}\else\textsc{Short--Long}\fi}
\renewcommand{\flat}{\ifmmode\text{\scshape Flat}\else\textsc{Flat}\fi}

\newcommand{\rankone}[1]{\cellcolor[RGB]{107,174,214}{#1}}
\newcommand{\ranktwo}[1]{\cellcolor[RGB]{158,202,225}{#1}}
\newcommand{\rankthree}[1]{\cellcolor[RGB]{198,219,239}{#1}}

\newcommand{\yes}{\textcolor{green!50!black}{\ensuremath{\checkmark}}}
\newcommand{\no}{\textcolor{red!70!black}{\ensuremath{\times}}}
\newtcolorbox{casecard}[1]{
  enhanced,
  breakable,
  colback=white,
  colframe=black,
  colbacktitle=black,
  coltitle=white,
  fonttitle=\bfseries\fontsize{9}{11}\selectfont,
  fontupper=\fontsize{9}{11}\selectfont,
  title=#1,
  boxrule=0.5pt,
  arc=1.5mm,
  boxsep=1mm,
  left=1.2mm,
  right=1.2mm,
  top=1.2mm,
  bottom=1.2mm
}

\title{\method: Policy Memory and Belief Revision for Language-Model Agents\\in Non-Stationary Pair Trading}
\author{
 \textbf{Polydoros Giannouris\textsuperscript{$\pmb\dagger$,*}},
 \textbf{Yuechen Jiang\textsuperscript{$\pmb\dagger$,*}},
 \textbf{Lingfei Qian\textsuperscript{$\clubsuit$}},\\
 \textbf{Yuyan Wang\textsuperscript{$\pmb\dagger$}},
 \textbf{Xueqing Peng\textsuperscript{$\clubsuit$}},
 \textbf{Jimin Huang\textsuperscript{$\pmb\dagger,\clubsuit$}},\\
 \textbf{Guojun Xiong\textsuperscript{$\spadesuit$}},
 \textbf{Sophia Ananiadou\textsuperscript{$\pmb\dagger,\diamondsuit$}}
\\
 \textsuperscript{$\pmb\dagger$}The University of Manchester, UK\\
 \textsuperscript{$\clubsuit$}The Fin AI, USA\\
 \textsuperscript{$\spadesuit$}Harvard University, USA\\
 \textsuperscript{$\diamondsuit$}Archimedes/Athena RC, Greece\\
 \texttt{\small{
 \{polydoros.giannouris,yuyan.wang-2,sophia.ananiadou\}@manchester.ac.uk}}
 \\
 \texttt{\small{
 \{yuechen.jiang,jimin.huang\}@postgrad.manchester.ac.uk}}
 \\
 \texttt{\small{
 \{lfqian94,xueqing.peng2024,xionggj1\}@gmail.com}}
}

\begin{document}
\maketitle
\begingroup
\renewcommand{\thefootnote}{*}
\footnotetext{Equal Contribution}
\endgroup
\begin{abstract}
Can an LLM agent turn experience into reusable policy while also revising its failing beliefs? We investigate this question with \method{}, a memory framework based on experience compression, validation, and policy abstraction, distinguishing reusable action guidance from the belief supporting those actions. 
We evaluate 11 models in 13 backbone configurations under Beta and No-Beta exposures against six baseline policies from four families. In the 75-session main evaluation, the median-return runs achieve a highest cumulative return of 31.2\% with Qwen3-32B thinking under Beta and a highest Sharpe ratio of 4.57 with Gemini 3.1 Flash-Lite under Beta. Most configurations outperform passive strategies, although benefits depend on the backbone and exposure setting. Controlled ablations show that policy consolidation and belief revision provide complementary benefits, while longer-horizon and cross-regime experiments indicate that these benefits can persist under changing conditions. More updates or pair switches do not consistently improve returns. 
These findings suggest that continual adaptation benefits from both validated policy abstraction and reassessment of the underlying belief. Code 
\footnote{\url{https://github.com/Carolyn-Jiang/MOIRA}}
and data\footnote{\href{https://huggingface.co/datasets/TheFinAI/Herculean}{Herculean dataset}, released by TheFinAI under CC BY 4.0} are available.
\end{abstract}

\section{Introduction}

\begin{figure}[t]
\centering
\includegraphics[width=0.98\columnwidth]{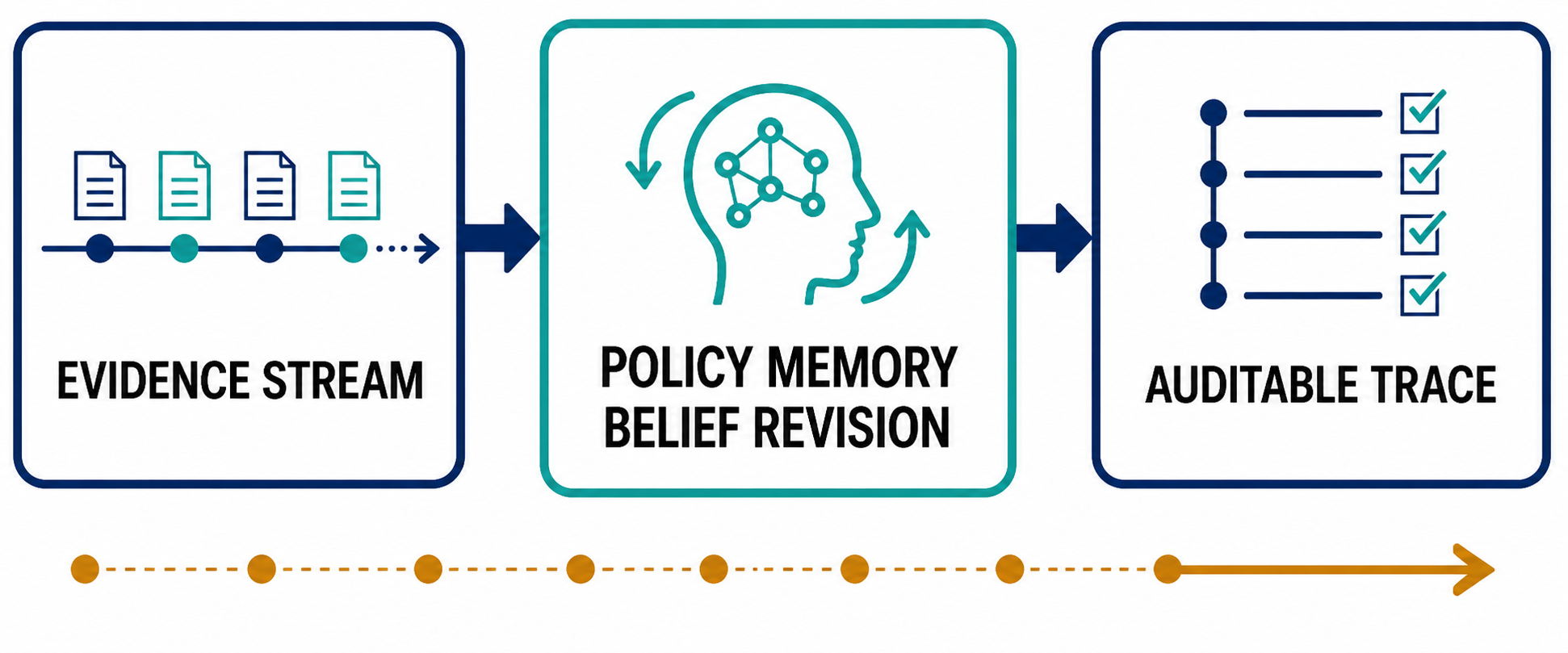}
\caption{Conceptual task of \method{}}
\label{fig:concept}
\end{figure}

Adaptive intelligence requires memory that turns experience into reusable knowledge, from classical consolidation and experience replay to reflective language agents \citep{mcclelland1995complementary,reyna1995fuzzy,sutton1991dyna,shinn2023reflexion,zhao2024expel}. Yet retrieval and terminal success give little insight into whether or not an agent learned a reusable policy or revised its latent hypothesis when evidence contradicted it \citep{maharana2024locomo,wu2025longmemeval,tan2025membench,ma2024agentboard}. We therefore ask: \emph{Can an LLM agent turn experience into reusable policy and revise failing beliefs?} Figure~\ref{fig:concept} abstracts the evaluation problem: observations enter a latent adaptive state, actions produce delayed factual and counterfactual feedback, and the output is an auditable policy-and-belief trajectory. This requires a setting with consequential feedback, non-stationarity, and independently observable update paths; pair trading supplies a compact financial instantiation through delayed returns and changing relative-price relationships.

\begin{table*}[!t]
\centering
\scriptsize
\setlength{\tabcolsep}{2.5pt}
\begin{adjustbox}{width=\textwidth}
\begin{tabular}{lcccccc}
\toprule
\textbf{Evaluation or method} & \shortstack{\textbf{Raw Data}\\\textbf{Memory}} & \shortstack{\textbf{Consequential}\\\textbf{actions}} & \shortstack{\textbf{Policy}\\\textbf{abstraction}} & \shortstack{\textbf{Counterfactual}\\\textbf{Reflection}} & \shortstack{\textbf{Belief}\\\textbf{update}} & \shortstack{\textbf{Belief}\\\textbf{revision}} \\
\midrule
LoCoMo \citep{maharana2024locomo} & \yes & \no & \no & \no & \no & \no \\
LongMemEval \citep{wu2025longmemeval} & \yes & \no & \no & \no & \no & \no \\
MemBench \citep{tan2025membench} & \yes & \no & \no & \no & \no & \no \\
Mem2ActBench \citep{shen2026mem2actbench} & \yes & \yes & \no & \no & \no & \no \\
AgentBench \citep{liu2024agentbench} & \yes & \yes & \no & \no & \no & \no \\
AgentBoard \citep{ma2024agentboard} & \yes & \yes & \no & \no & \no & \no \\
Reflexion \citep{shinn2023reflexion} & \no & \yes & \yes & \no & \no & \no \\
ExpeL \citep{zhao2024expel} & \yes & \yes & \yes & \no & \no & \no \\
\textsc{FinMem} \citep{yu2025finmem} & \yes & \yes & \yes & \no & \no & \no \\
FinAgent \citep{zhang2024finagent} & \yes & \yes & \yes & \no & \no & \no \\
FinCon \citep{yu2024fincon} & \yes & \yes & \yes & \no & \yes & \no \\
\method{} (ours) & \no & \yes & \yes & \yes & \yes & \yes \\
\bottomrule
\end{tabular}
\end{adjustbox}
\caption{Memory and adaptation capabilities. Raw Data Memory persistently stores raw interactions or observations; \method{} retains only derived state. Belief Update edits the system prompt; Belief Revision reconstructs the active-pair relationship. \yes{}/\no{}: presence/absence.}
\label{tab:comparison}
\end{table*}

Prior work addresses complementary parts of this question. LoCoMo \citep{maharana2024locomo}, LongMemEval \citep{wu2025longmemeval}, and MemBench \citep{tan2025membench} evaluate recall and updating, while Mem2ActBench \citep{shen2026mem2actbench}, AgentBench \citep{liu2024agentbench}, and AgentBoard \citep{ma2024agentboard} extend evaluation toward consequential interaction. Reflexion \citep{shinn2023reflexion} and ExpeL \citep{zhao2024expel} extract linguistic experience; \textsc{FinMem} \citep{yu2025finmem}, FinAgent \citep{zhang2024finagent}, and \textsc{FinCon} \citep{yu2024fincon} instantiate memory and feedback in finance. AgeMem learns joint memory management \citep{yu2026agentic}, and KTD-Fin controls historical market information and attributes trading returns \citep{zhu2026knowingdoing}. A remaining gap is the joint coordination of validated policy consolidation and daily correction with replacement of a failing relationship within one evolving task. Table~\ref{tab:comparison} summarizes the relevant operational roles.

We introduce \method{}, a traceable state-update architecture for LLM agents in non-stationary pair trading. The main contribution is to build an audit trail that separates available resources from training/warmup, daily Counterfactual Reflection, replay-gated global policy updates, and lifecycle-driven pair revision. Short-term memory retains Counterfactual Reflection from the three most recent trading days, while long-term memory provides validated guidance for pair selection (Appendix~\ref{app:architecture}). The design uses fixed model parameters and compact derived memory, without a layered raw-data archive, making both state transitions explicit and auditable.

The evaluations reveal \textbf{backbone-dependent adaptation}: most configurations exceed passive baselines, and larger open-weight models generally perform better, while explicit reasoning provides no consistent advantage (Table~\ref{tab:main}). They show \textbf{exposure-conditioned feedback}, as beta hedging (Beta) helps some backbones and harms others, and \textbf{validation over update volume}, since more updates or switches need not yield higher returns (Appendix~\ref{app:case}). Leading configurations exceed \textsc{FinMem} in return with lower long-horizon test-token use, although \textsc{FinMem} retains strong risk control. Longer-horizon and cross-regime evaluations support conditional persistence and adaptability. Ablations reveal distinct degradations when policy consolidation or reselection is disabled, supporting their complementary roles (Tables~\ref{tab:a1}--\ref{tab:a4}). Our contributions are:
\begin{itemize}
    \item We design an auditable dual-timescale state-update workflow for LLM agents in non-stationary settings, separating shared input resources from training/warmup and daily reflection: replay validation gates global policy updates, while lifecycle evidence gates revision of the active pair relationship.
    \item We evaluate this design in pair trading, a consequential partially observed environment, against passive, rule-based, reinforcement-learning, and LLM-agent baselines across backbones and exposure settings.
    \item We find that policy consolidation and lifecycle-driven pair-relationship revision have complementary, conditional effects: removing either produces a distinct degradation, while return--risk profiles vary across backbones and exposure settings. The compact derived state also uses fewer long-horizon test tokens than \textsc{FinMem}.
\end{itemize}

\section{Related Work}

\textbf{Memory representation and use.} Agent memories range from retrieved episodes and tiered context \citep{park2023generative,zhong2024memorybank,packer2023memgpt} to editable graphs, hierarchical stores, and evolving memory networks \citep{wang2024emgrag,kang2025memoryos,xu2025amem,latimer2026hindsight}. LoCoMo, LongMemEval, MemBench, and Mem2ActBench evaluate recall, updating, memory operations, and tool-conditioned use \citep{maharana2024locomo,wu2025longmemeval,tan2025membench,shen2026mem2actbench}. AgeMem jointly learns long- and short-term memory management through reinforcement learning \citep{yu2026agentic}. Related surveys and evaluations further study experience abstraction and reflective memory \citep{luo2026memorysurvey,tan2025reflective}.

\textbf{Learning behavior from linguistic feedback.} ReAct, Reflexion, and Self-Refine interleave action or revision with language feedback \citep{yao2023react,shinn2023reflexion,madaan2023selfrefine}; ExpeL and MetaReflection extract cross-task principles \citep{zhao2024expel,gupta2024metareflection}; and Voyager and Agent Workflow Memory retain skills or workflows \citep{wang2024voyager,wang2025awm}. Continuous and fine-grained evaluations examine improvement streams and memory-operation credit \citep{wu2024streambench,ma2026finemem,qin2026implicit}. Financial agents use layered memory, verbal reinforcement, tools, or role coordination \citep{yu2025finmem,zhang2024finagent,yu2024fincon,xiao2024tradingagents}; FinCon additionally derives belief updates from comparisons between episode trajectories \citep{yu2024fincon}. Related reinforcement-learning work studies learning from simulated or relabeled experience \citep{sutton1991dyna,andrychowicz2017her,sutton2018rl}.

\textbf{Sequential agent evaluation.} AgentBench, GAIA, ToolLLM, WebArena, Mind2Web, OSWorld, and AgentBoard evaluate interaction, planning, and tool use \citep{liu2024agentbench,mialon2024gaia,qin2024toolllm,zhou2024webarena,deng2023mind2web,xie2024osworld,ma2024agentboard}; PIXIU and FinBen cover financial language capabilities \citep{xie2023pixiu,xie2024finben}. KTD-Fin masks stock identifiers and calendar information to control exposure to memorized market information, and attributes returns to market, style, and stock-selection components \citep{zhu2026knowingdoing}.

\section{Method}
\label{sec:method}


\begin{figure*}[!t]
\centering
\includegraphics[width=\textwidth]{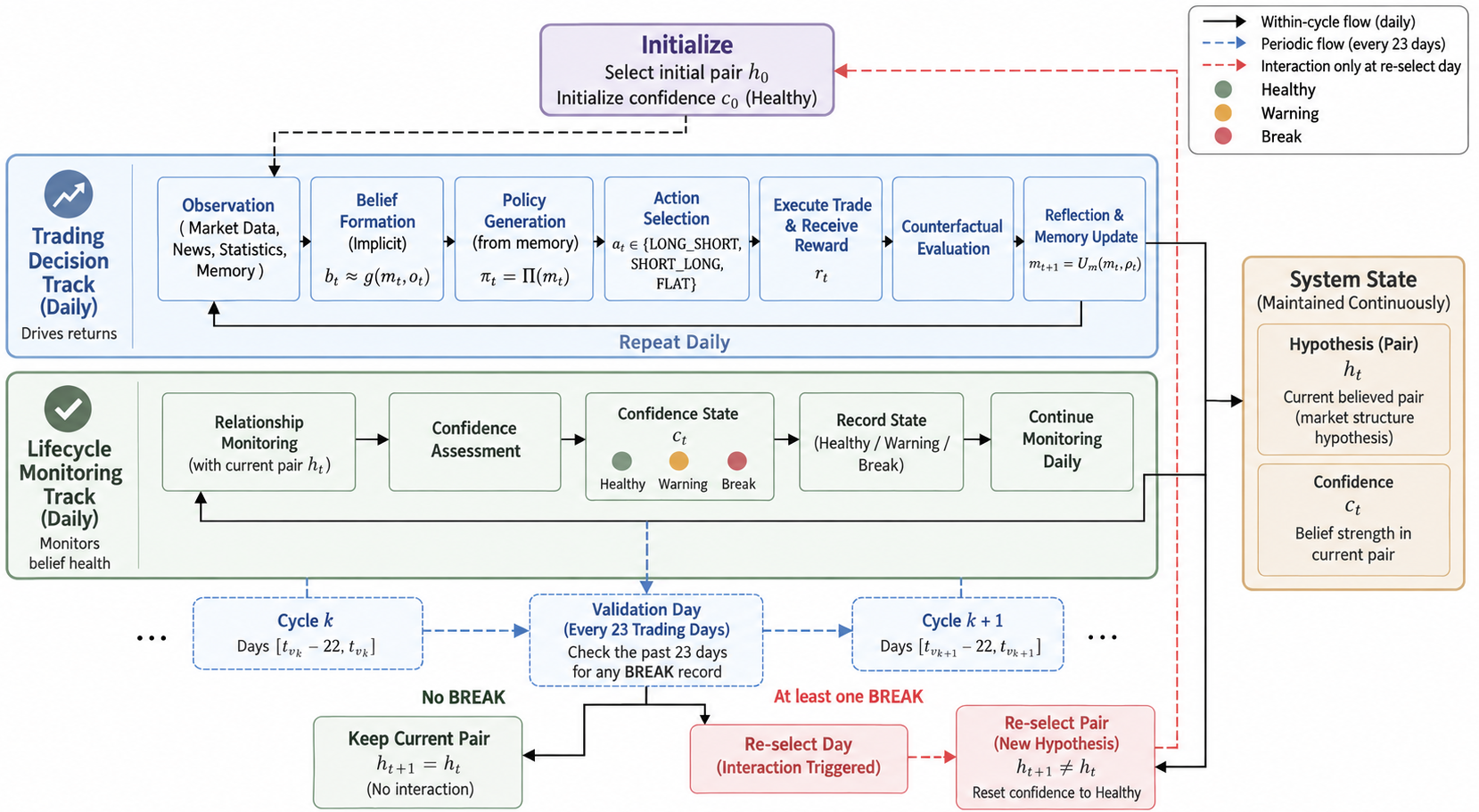}
\caption{Overall \method{} workflow.}
\label{fig:workflow}
\end{figure*}

\subsection{Pair Trading Task Formulation}
We formulate pair trading as a finite partially observable Markov decision process \citep{kaelbling1998pomdp}
\begin{equation}
\begin{aligned}
\mathcal{M}&=(\mathcal{S},\mathcal{O},\mathcal{A},\mathcal{T},\Omega,\mathcal{R}),\\
\mathcal{A}&=\{\mathrm{LS},\mathrm{SL},\mathrm{F}\}.
\end{aligned}
\end{equation}
Here LS, SL, and F denote the potential daily actions; Long--Short, Short--Long, and Flat. The latent state $\sigma_t\in\mathcal{S}$ contains the market regime and the validity of relative-price relationships, while $o_t\in\mathcal{O}$ contains only prices, derived statistics, prior-session news, and memory available by session $t$. The agent maintains an active pair hypothesis $h_t=(i_t,j_t)\in\mathcal{H}_t$ and chooses $a_t\in\mathcal{A}$. Let $g_t(h_t)$ be the one-session return, realized after the next prices arrive, of the deterministic long-$i_t$, short-$j_t$ position; then
\begin{equation}
\begin{aligned}
r_t(a_t,h_t)=
\begin{cases}
g_t(h_t),&a_t=\mathrm{LS},\\
-g_t(h_t),&a_t=\mathrm{SL},\\
0,&a_t=\mathrm{F},
\end{cases}
\\[-1mm]
J(\pi)=\mathbb{E}_{\pi}\!\left[\sum_{t=1}^{T}r_t\right].
\end{aligned}
\end{equation}
Thus an action is evaluated against all three one-step outcomes under the same held hypothesis. Position sizing, hedge-ratio estimation, candidate screening, and all financial statistics are defined in Appendix~\ref{app:financial-statistics}.

\subsection{Architecture of \method}
\method{} maintains an explicit adaptive state $x_t=(h_t,c_t,m^G,m_t^D)$, where $c_t$ is lifecycle confidence, $m^G$ is validated global policy memory, and $m_t^D$ is rolling daily reflection memory. Its workflow is
\begin{equation}
\begin{aligned}
a_t&=\pi_\theta(o_t,x_t),\\
\xi_t&=G(o_t,a_t,r_t(\mathcal{A})),\\
(m_{t+1}^D,h_{t+1},c_{t+1})&=U(x_t,\xi_t,o_{t+1}),\\
\theta_{t+1}&=\theta_t.
\end{aligned}
\end{equation}
so experience changes explicit memory and belief state but never model parameters. Figure~\ref{fig:workflow} summarizes the two update paths; Appendix~\ref{app:architecture} provides implementation views, and Appendix~\ref{app:prompts} gives every model prompt.

\subsubsection{Policy Memory and Counterfactual Reflection}
After next-session prices arrive, $\xi_t$ records the chosen action, realized payoff, all-action alternatives, and deterministic error label; $m_{t+1}^D=U_D(m_t^D,\xi_t)$ retains the three most recent records. The replay objective $J(m^G)$ is cumulative training return, defined as the sum of daily returns, $\sum_t r_t$ (Appendix~\ref{app:adaptation_details}). A replay round proposes ticker-independent principles $\widetilde m^G_{k+1}$ and accepts them only under
\begin{equation}
m^G_{k+1}=\begin{cases}
\widetilde m^G_{k+1},&J(\widetilde m^G_{k+1})>J(m^G_k),\\
m^G_k,&\text{otherwise}.
\end{cases}
\end{equation}
Update-loop details are in Appendix~\ref{app:adaptation_details}. Testing freezes $m^G$ while continuing daily updates to $m_t^D$, separating validated policy consolidation from short-horizon correction. This memory path corresponds to Figure~\ref{fig:model-architecture-a} and relates verbal reflection to experience reuse without weight updates \citep{shinn2023reflexion,sutton1991dyna,sutton2018rl}.

\subsubsection{Pair Selection as Belief Formation}
Each eligible pair $h\in\mathcal{H}_t$ is a falsifiable hypothesis about a persistent relative-price relationship. All methods begin with the same rule-based statistical screening procedure, which uses correlation, cointegration, and stationarity diagnostics to construct $\mathcal{H}_t$. Baseline selectors choose from this candidate set using only the statistical indicators, although \textsc{FinMem} subsequently uses news and company context in its action and memory modules. \method{} instead treats the indicators as quantitative evidence and combines them with contemporaneous news and company business relationships when choosing the active pair:
\begin{equation}
h_t=\arg\max_{h\in\mathcal{H}_t}S(h\mid o_t,m^G),
\end{equation}
where $S$ is an implicit language-model score. Selection is attempted initially, and the 23-session scheduler retries only while no eligible pair is held. Figure~\ref{fig:model-architecture-b} shows the division between statistical screening and model selection; Appendix~\ref{app:financial-statistics} defines the tests and ranking.

\subsubsection{Lifecycle Monitoring and Belief Revision}
Let $s_{j,t}$ be the deterioration points assigned by component $j$ and $s_t$ their aggregate. The statistical lifecycle label is
\begin{equation}
\begin{aligned}
s_t&=\sum_{j=1}^{6}s_{j,t},\\[-1mm]
q_t&=
\begin{cases}
\textsc{Healthy},&s_t\leq3,\\
\textsc{Warning},&3<s_t<5,\\
\textsc{Break},&s_t\geq5.
\end{cases}
\end{aligned}
\end{equation}
With severity order \textsc{Healthy}$\prec$\textsc{Warning}$\prec$\textsc{Break}, the independent news-relation label $c_t^{\mathrm{news}}$ can only raise the final state:
\begin{equation}
c_t=\max_{\prec}\{q_t,c_t^{\mathrm{news}}\}.
\end{equation}
Pair replacement is not triggered by an unconstrained language-model decision. The lifecycle module aggregates statistical deterioration, relevant news, and realized trading evidence. Only after the accumulated evidence reaches the predefined \textsc{Break} condition can \method{} reconsider the active hypothesis at the next validation boundary. A \textsc{Break} records incompatibility with $h_t$ but never forces an immediate switch. At each 23-session boundary $t_v$, let $\mathcal{W}_v=\{t_v-22,\ldots,t_v\}$ be the preceding cycle and define
\begin{equation}
B_v=\max_{\tau\in\mathcal{W}_v}
\mathbb{1}[c_\tau=\textsc{Break}].
\end{equation}
The replacement rule is
\begin{equation}
\label{eq:pair-reselection}
h_{t_v+1}=
\begin{cases}
\operatorname{Select}(\mathcal{H}_{t_v+1}),&B_v=1,\\
h_{t_v},&B_v=0.
\end{cases}
\end{equation}
Thus the 23-session scheduler cannot replace an active pair; Equation~\ref{eq:pair-reselection} is the only replacement rule once a pair is held. This isolates daily policy adaptation from slower belief revision. Figure~\ref{fig:model-architecture-c} provides the implementation view, while Appendix~\ref{app:financial-statistics} defines all six score components, threshold sources, and validation hyperparameters.

\section{Experiments}

\subsection{Experiment Setup}
We perform three independent runs per setting and report the median-return run in tables and plots. No-Beta uses equal-magnitude share legs; Beta uses a rolling 20-session hedge ratio (Appendix~\ref{app:financial-statistics}). We evaluate 11 model identifiers in 13 backbone configurations under both exposures: OpenAI's GPT-4.1-mini (G41) \citep{openai2025gpt41}, GPT-5.4-mini (G54) and GPT-5.4-nano (G54n) \citep{openai2026gpt54small,openai2026gpt54card}, and o4-mini \citep{openai2025o4mini}; DeepSeek-V4.1-Flash (DS) \citep{deepseek2026v4}; Google DeepMind's Gemini 3.1 Flash-Lite (GF) and Gemini 3.8 Flash (GL), targeting efficient inference and higher-capability agentic reasoning, respectively \citep{geminiteam2023gemini,google2026gemini31flashlite,google2026gemini38flash}; and open-weight Llama-3.1-8B/70B-Instruct (L8/L70) \citep{meta2024llama31} and Qwen3-4B/32B (Q4/Q32), each Qwen model with thinking on (thinking) or off (no-think) \citep{qwen2025qwen34b,qwen2025qwen332b}.

The six baseline policies span four families: passive fixed-direction \ls{}, \sl{}, and Equal-Weight; rule-based Engle--Granger with Z-score thresholds (EG+Z) \citep{engle1987cointegration}; reinforcement-learning Engle--Granger with a Double Deep Q-Network (EG-DDQN) \citep{engle1987cointegration,mnih2015human,vanhasselt2016deep}; and \textsc{FinMem} with GPT-4.1-mini \citep{yu2025finmem}. All methods share rule-based statistical screening and candidate construction. Only \method{} combines statistical indicators, news, and company business relationships during pair selection; \textsc{FinMem} uses news and business context after selection. Appendix Table~\ref{tab:mechanism-comparison} distinguishes available resources from training, reflection, and update workflows. Price-only comparisons assess the full system, including its additional information. Policy-update and reselection ablations retain these sources while disabling the corresponding mechanism (Tables~\ref{tab:a3}--\ref{tab:a4}).

All main-experiment backbones share a ten-stock pool, 27 training sessions (2 January--11 February 2025), and 75 test sessions (12 February--30 May 2025); selection uses prices from the causal ten-year window ending on each selection date (Appendix~\ref{app:data}). Both Gemini models use low thinking effort and a 4,096-token output limit; Llama/Qwen use a 16,384-token limit (Appendix~\ref{app:models}). Pair replacement is checked only at 23-session validation boundaries against the preceding cycle's break evidence. Appendix~\ref{app:models} lists inference settings; Appendices~\ref{app:prompts}--\ref{app:baselines} specify prompts, data, model identifiers, and controls; Appendix~\ref{app:metrics} defines metrics.

\subsection{Main Results}

\begin{figure*}[!t]
\centering
\includegraphics[width=\textwidth]{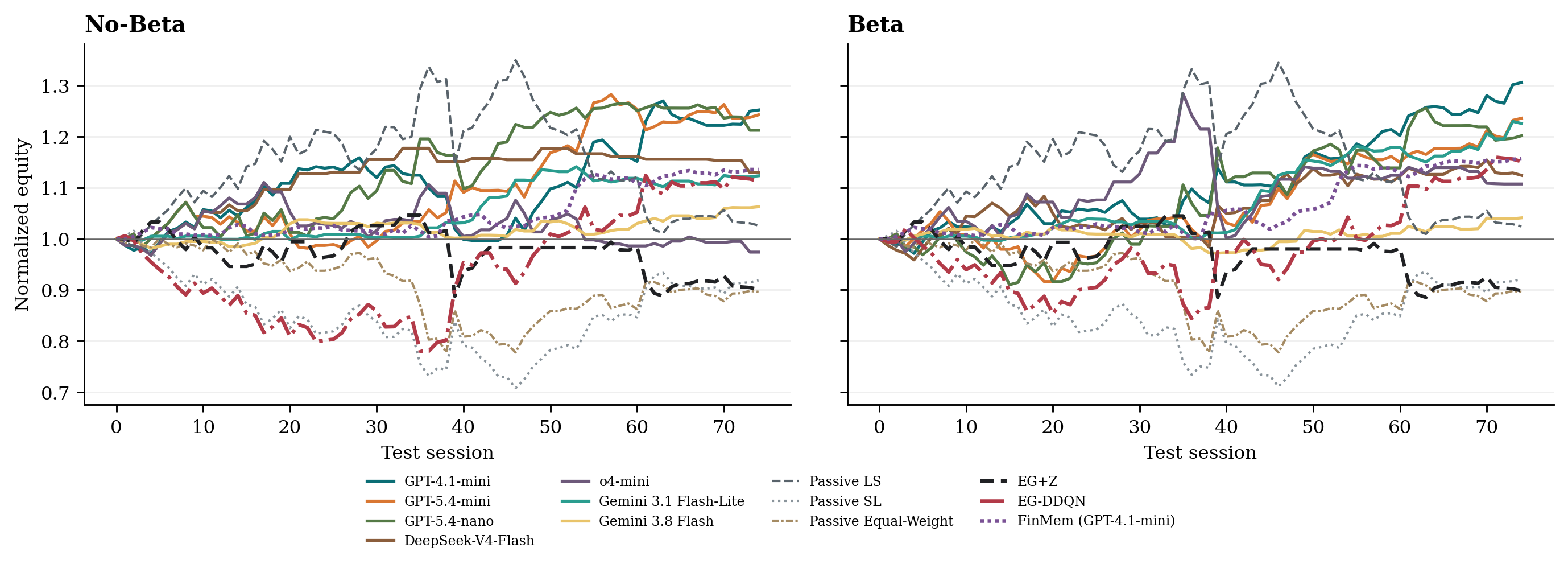}
\caption{No-Beta and Beta cumulative returns for seven \method{} backbones and passive, EG+Z, EG-DDQN, and \textsc{FinMem} baselines. Endpoints match Table~\ref{tab:main}; Appendix~\ref{app:additional-results} covers all 13 configurations.}
\label{fig:returns-main}
\par
{
\centering
\fontsize{9}{10.5}\selectfont
\setlength{\tabcolsep}{0.5pt}
\begingroup
\setlength{\arrayrulewidth}{0.25pt}
\resizebox{\textwidth}{!}{%
\begin{tabular}{@{}ll!{\tablegroupdash}*{6}{r}!{\tablegroupdash}*{7}{r}!{\tablegroupdash}*{3}{r}!{\tablegroupdash}*{2}{r}!{\tablegroupdash}r@{}}
\toprule
\multicolumn{2}{c!{\tablegroupdash}}{} & \multicolumn{6}{c!{\tablegroupdash}}{\textbf{Closed-source \method{}}} & \multicolumn{7}{c!{\tablegroupdash}}{\textbf{Open-weight \method{}}} & \multicolumn{3}{c!{\tablegroupdash}}{\textbf{Passive}} & \multicolumn{2}{c!{\tablegroupdash}}{\textbf{Traditional}} & \textbf{\textsc{FinMem}}\\
\midrule
\textbf{Exp.} & \textbf{Met.} & G41 & G54 & G54n & o4 & GF & GL & DS & L8 & L70 & Q4N & Q4T & Q32N & Q32T & LS & SL & EW & EZ & DQ & G41\\
\midrule
B & Ret. & \ranktwo{30.6} & 23.6 & 20.2 & 10.7 & 22.6 & 4.1 & 12.4 & 12.4 & 20.6 & \textminus6.3 & 2.1 & 15.2 & \rankone{31.2} & 2.4 & \textminus7.9 & \textminus10.4 & \textminus10.17 & 15.10 & 15.64\\
 & Sh. & \ranktwo{3.62} & 2.58 & 1.62 & 1.04 & \rankone{4.57} & 1.17 & 1.57 & 1.53 & 2.20 & \textminus0.60 & 0.33 & 1.31 & 2.36 & 0.36 & \textminus0.46 & \textminus0.86 & \textminus0.87 & 1.27 & 2.91\\
 & MDD & \textminus4.8 & \textminus13.0 & \textminus10.9 & \textminus22.0 & \textminus3.7 & \textminus5.0 & \textminus9.2 & \textminus19.0 & \textminus17.7 & \textminus19.0 & \textminus19.8 & \textminus19.6 & \textminus14.8 & \textminus24.9 & \textminus30.0 & \textminus22.3 & \textminus15.26 & \textminus17.21 & \textminus3.71\\
\midrule
NB & Ret. & \rankthree{25.2} & 24.3 & 21.2 & \textminus2.6 & 12.3 & 6.3 & 12.9 & 4.2 & 2.0 & 4.8 & \textminus4.1 & 15.8 & 15.4 & 2.6 & \textminus8.1 & \textminus10.4 & \textminus9.92 & 11.25 & 13.53\\
 & Sh. & 2.66 & 2.79 & 2.16 & \textminus0.28 & \rankthree{3.25} & 1.94 & 2.62 & 0.54 & 0.29 & 0.58 & \textminus0.17 & 1.56 & 2.07 & 0.37 & \textminus0.47 & \textminus0.86 & \textminus0.84 & 0.98 & 3.07\\
 & MDD & \textminus14.0 & \textminus8.4 & \textminus8.1 & \textminus12.3 & \ranktwo{\textminus3.4} & \rankthree{\textminus3.5} & \textminus4.1 & \textminus18.2 & \textminus16.6 & \textminus16.0 & \textminus20.1 & \textminus9.8 & \textminus7.1 & \textminus25.1 & \textminus30.3 & \textminus22.3 & \textminus15.17 & \textminus22.42 & \rankone{\textminus2.54}\\
\bottomrule
\end{tabular}
}

\endgroup
\captionof{table}{Main results for 13 backbone configurations. B/NB: Beta/No-Beta. Blue marks the top three across exposures. Ret.: return (\%); Sh.: Sharpe ratio; MDD: maximum drawdown (\%). EW: Equal-Weight; G41/G54/G54n: GPT-4.1-mini/GPT-5.4-mini/GPT-5.4-nano; DS: DeepSeek-V4.1-Flash; o4: o4-mini; GF/GL: Gemini 3.1 Flash-Lite/3.8 Flash; L8/L70: Llama-3.1-8B-Instruct/70B-Instruct; Q4N/Q4T: Qwen3-4B no-think/thinking; Q32N/Q32T: Qwen3-32B no-think/thinking. These 13 columns use \method{}. LS/SL/EW: passive policies; EZ: rule-based Engle--Granger with z-score thresholds; DQ: Engle--Granger Double Deep Q-Network; The G41 column under \textsc{FinMem} uses GPT-4.1-mini.}
\label{tab:main}
}
\end{figure*}

Most backbones outperform passive strategies in return: 10 of 13 configurations under No-Beta and 11 under Beta (Table~\ref{tab:main}). All exceed EG+Z, while seven exceed EG-DDQN under each exposure. Q32T leads Beta return, narrowly ahead of GPT-4.1-mini, whereas GF leads Sharpe. 
\textbf{Scale generally helps open-weight models.} Q32 exceeds 4B in return and Sharpe across both reasoning modes and exposures. L70 likewise exceeds 8B under Beta, but not under No-Beta. Among proprietary models, GPT-5.4-mini exceeds the smaller nano in both metrics, whereas GF exceeds GL despite its efficiency-oriented positioning. 

\textbf{Reasoning and exposure interact.} The reasoning-oriented o4-mini underperforms GPT-4.1-mini in return and Sharpe under both exposures. This does not extend uniformly to Qwen: thinking improves Beta performance at both sizes, but under No-Beta lowers 4B return and leaves 32B return nearly unchanged despite higher Sharpe. Overall, Beta favors seven configurations and No-Beta six. Higher returns can also accompany fewer updates or switches, distinguishing local replay validation from subsequent transfer (Appendix~\ref{app:case}).
\textbf{Memory design trades return against risk.} Leading \method{} configurations exceed \textsc{FinMem} in return, while \textsc{FinMem} retains the best No-Beta drawdown. GF and GPT-4.1-mini Beta also exceed its Sharpe. Compact validated state achieves these gains with about \textbf{75\% fewer} GPT-4.1-mini A1 test tokens (Appendix~\ref{app:token_usage}).

\subsection{Ablation Studies}

\subsubsection{A1: Long-Horizon Evaluation}
A1 evaluates GPT-4.1-mini with Beta exposure (G41-B) and GPT-5.4-mini with No-Beta exposure (G54-N); A2--A4 use the same configurations (Appendix~\ref{app:baselines}). Training covers 28 sessions (2 December 2024--13 January 2025), and testing covers 304 sessions (14 January 2025--31 March 2026), yielding 303 finalized transitions (Appendix~\ref{app:data}).

\begin{figure}[!ht]
\centering
\includegraphics[width=\columnwidth]{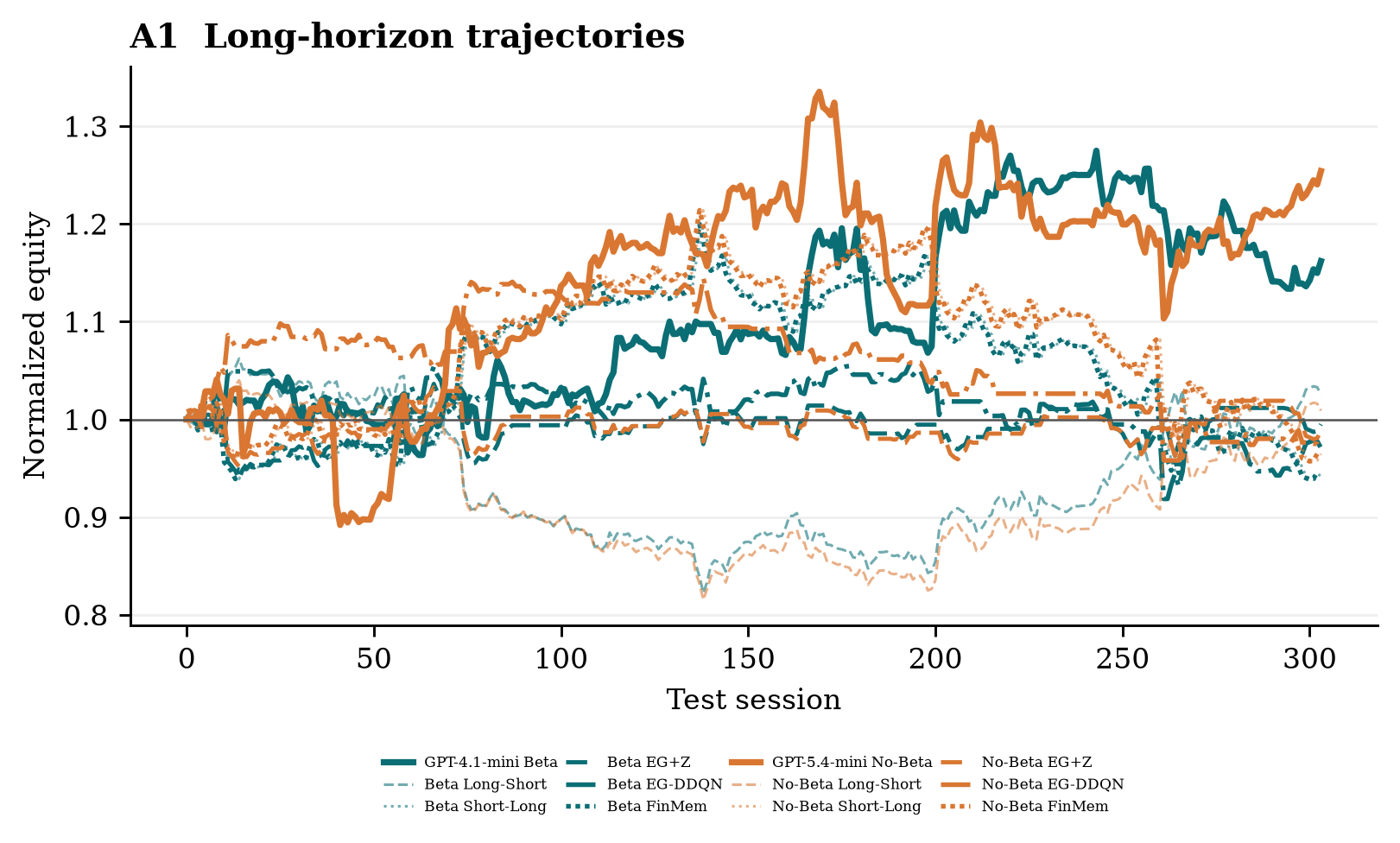}
\caption{A1 trajectories for \method{} and matched baselines.}
\label{fig:ablation-a1}
\par
{
\centering
\fontsize{9}{10.5}\selectfont
\setlength{\tabcolsep}{0.7pt}
\begin{tabular}{@{}l*{6}{r}@{}}
\toprule
 & \multicolumn{6}{c}{\textbf{Beta}}\\
\cmidrule(lr){2-7}
\textbf{Policy} & \textbf{\method{}} & \textbf{LS} & \textbf{SL} & \textbf{EG+Z} & \textbf{DDQN} & \textbf{FM}\\
Backbone & G41 & -- & -- & -- & -- & G41\\
\midrule
Ret. & \ranktwo{16.19} & \rankthree{2.56} & $-5.42$ & $-0.50$ & $-2.92$ & $-5.42$\\
Sh. & \ranktwo{0.61} & \rankthree{0.08} & $-0.35$ & $-0.18$ & $-0.25$ & $-0.35$\\
MDD & \rankthree{$-11.1$} & $-22.6$ & $-21.8$ & \ranktwo{$-7.87$} & $-12.97$ & $-21.79$\\
\bottomrule
\addlinespace[4pt]
\toprule
 & \multicolumn{6}{c}{\textbf{No-Beta}}\\
\cmidrule(lr){2-7}
\textbf{Policy} & \textbf{\method{}} & \textbf{LS} & \textbf{SL} & \textbf{EG+Z} & \textbf{DDQN} & \textbf{FM}\\
Backbone & G54 & -- & -- & -- & -- & G41\\
\midrule
Ret. & \rankone{25.41} & 0.90 & $-3.54$ & $-1.24$ & $-3.18$ & $-3.54$\\
Sh. & \rankone{0.83} & $-0.02$ & $-0.27$ & $-0.25$ & $-0.36$ & $-0.27$\\
MDD & $-17.4$ & $-21.6$ & $-21.2$ & \rankone{$-7.13$} & $-15.32$ & $-21.17$\\
\bottomrule
\end{tabular}
\captionof{table}{A1 long-horizon results. Blue marks the top three across exposures. G41/G54 denote GPT-4.1-mini/GPT-5.4-mini. DDQN: EG-DDQN; FM: \textsc{FinMem}; LS/SL: passive policies.}
\label{tab:a1}
}
\end{figure}

Both \method{} settings exceed directional and learned baselines over the longer horizon without test-time retraining (Table~\ref{tab:a1}; Figure~\ref{fig:ablation-a1}). GPT-5.4-mini No-Beta earns more than GPT-4.1-mini Beta (25.41\% versus 16.19\%), but incurs a deeper drawdown ($-17.4\%$ versus $-11.1\%$). Their shared advantage supports persistent adaptation through consolidated guidance, online reflection, and relationship maintenance, with a backbone--exposure trade-off in risk.

\subsubsection{A2: Cross-Regime Evaluation}

\begin{figure}[!ht]
\centering
\includegraphics[width=\columnwidth]{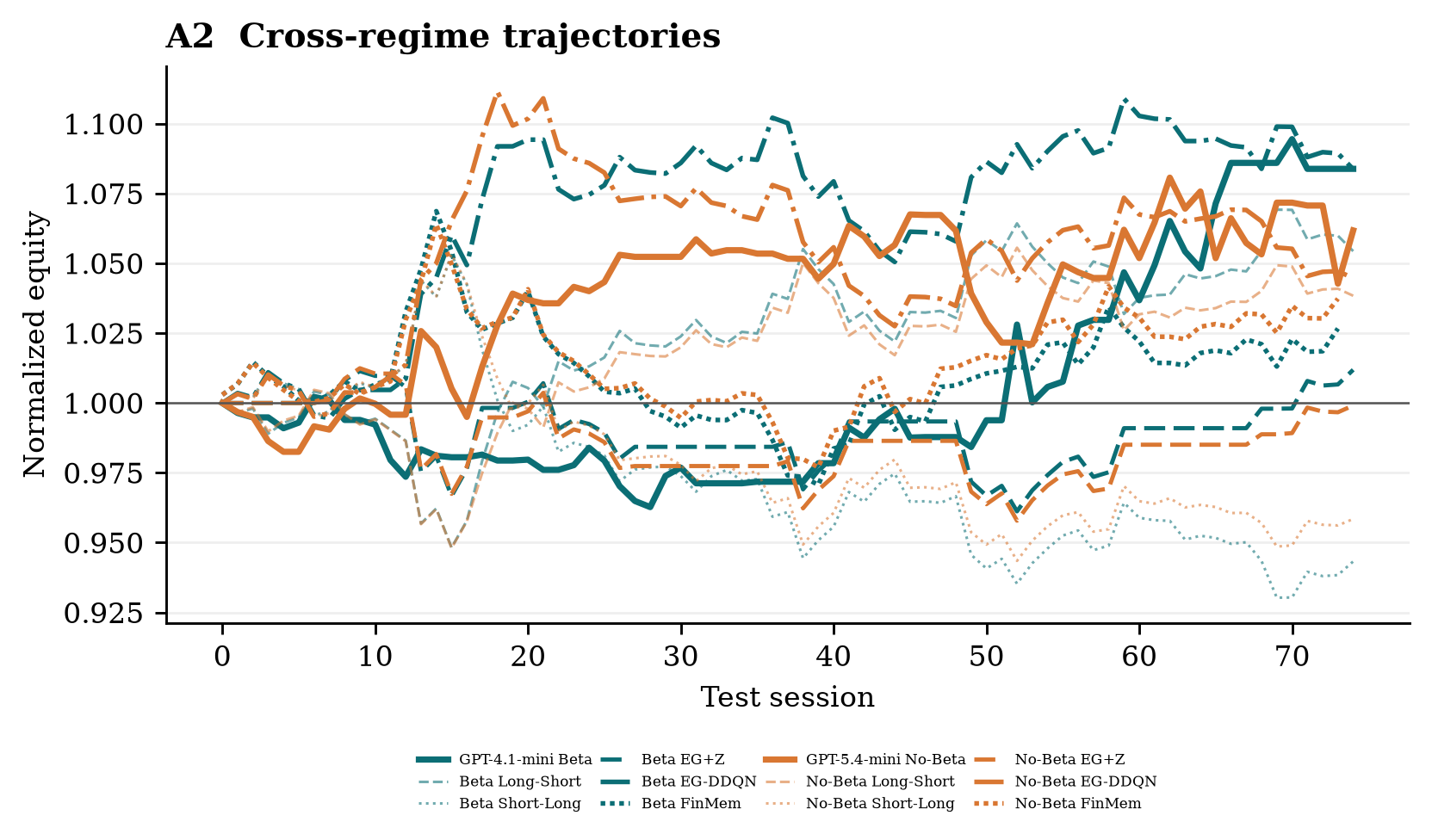}
\caption{A2 cross-regime trajectories for \method{} and matched baselines.}
\label{fig:ablation-a2}
\par
{
\centering
\fontsize{9}{10.5}\selectfont
\setlength{\tabcolsep}{0.7pt}
\begin{tabular}{@{}l*{6}{r}@{}}
\toprule
 & \multicolumn{6}{c}{\textbf{Beta}}\\
\cmidrule(lr){2-7}
\textbf{Policy} & \textbf{\method{}} & \textbf{LS} & \textbf{SL} & \textbf{EG+Z} & \textbf{DDQN} & \textbf{FM}\\
Backbone & G41 & -- & -- & -- & -- & G41\\
\midrule
Ret. & \rankone{8.39} & 5.44 & $-5.67$ & 1.19 & \ranktwo{8.37} & 2.67\\
Sh. & \ranktwo{1.93} & 1.23 & $-1.55$ & 0.22 & \rankone{2.07} & 0.62\\
MDD & \ranktwo{$-3.9$} & $-5.6$ & $-11.7$ & $-4.96$ & \rankone{$-3.86$} & $-9.17$\\
\bottomrule
\addlinespace[4pt]
\toprule
 & \multicolumn{6}{c}{\textbf{No-Beta}}\\
\cmidrule(lr){2-7}
\textbf{Policy} & \textbf{\method{}} & \textbf{LS} & \textbf{SL} & \textbf{EG+Z} & \textbf{DDQN} & \textbf{FM}\\
Backbone & G54 & -- & -- & -- & -- & G41\\
\midrule
Ret. & \rankthree{6.18} & 3.84 & $-4.15$ & $-0.09$ & 4.47 & 3.74\\
Sh. & \rankthree{1.27} & 0.90 & $-1.24$ & $-0.16$ & 1.07 & 0.95\\
MDD & \rankthree{$-4.4$} & $-5.6$ & $-10.4$ & $-5.39$ & $-6.54$ & $-8.18$\\
\bottomrule
\end{tabular}
\captionof{table}{A2 cross-regime results. Blue marks the top three across exposures. DDQN: EG-DDQN; FM: \textsc{FinMem}; LS/SL: passive policies.}
\label{tab:a2}
}
\end{figure}

A1 tests long-horizon persistence; A2 tests adaptation after retraining in a different market regime (Appendix~\ref{app:data}). A2 retrains policy memory on 28 sessions (2 June--11 July 2025) and tests in a bullish regime over 75 sessions (14 July--27 October 2025). GPT-4.1-mini Beta narrowly exceeds EG-DDQN in return (8.39\% versus 8.37\%) but trails its Sharpe and drawdown control; GPT-5.4-mini No-Beta exceeds its controls on all three metrics (Table~\ref{tab:a2}; Figure~\ref{fig:ablation-a2}). This contrast supports \textbf{cross-regime adaptability of the compression-and-validation procedure}, with backbone-dependent return--risk benefits.

\subsubsection{A3: Removing Global Policy Update}

A3 uses the main experiment's training and test windows and compares global policy-memory consolidation (belief update) with warmup-only reflection memory; A4 instead controls pair-hypothesis revision (Appendix~\ref{app:adaptation_details}). Both conditions share a single-round warmup and its initial reflection memory, and retain daily Counterfactual Reflection and lifecycle-controlled reselection; only post-warmup policy consolidation is enabled or disabled (Appendix~\ref{app:adaptation_details}).

\begin{figure}[!ht]
\centering
\includegraphics[width=0.85\columnwidth]{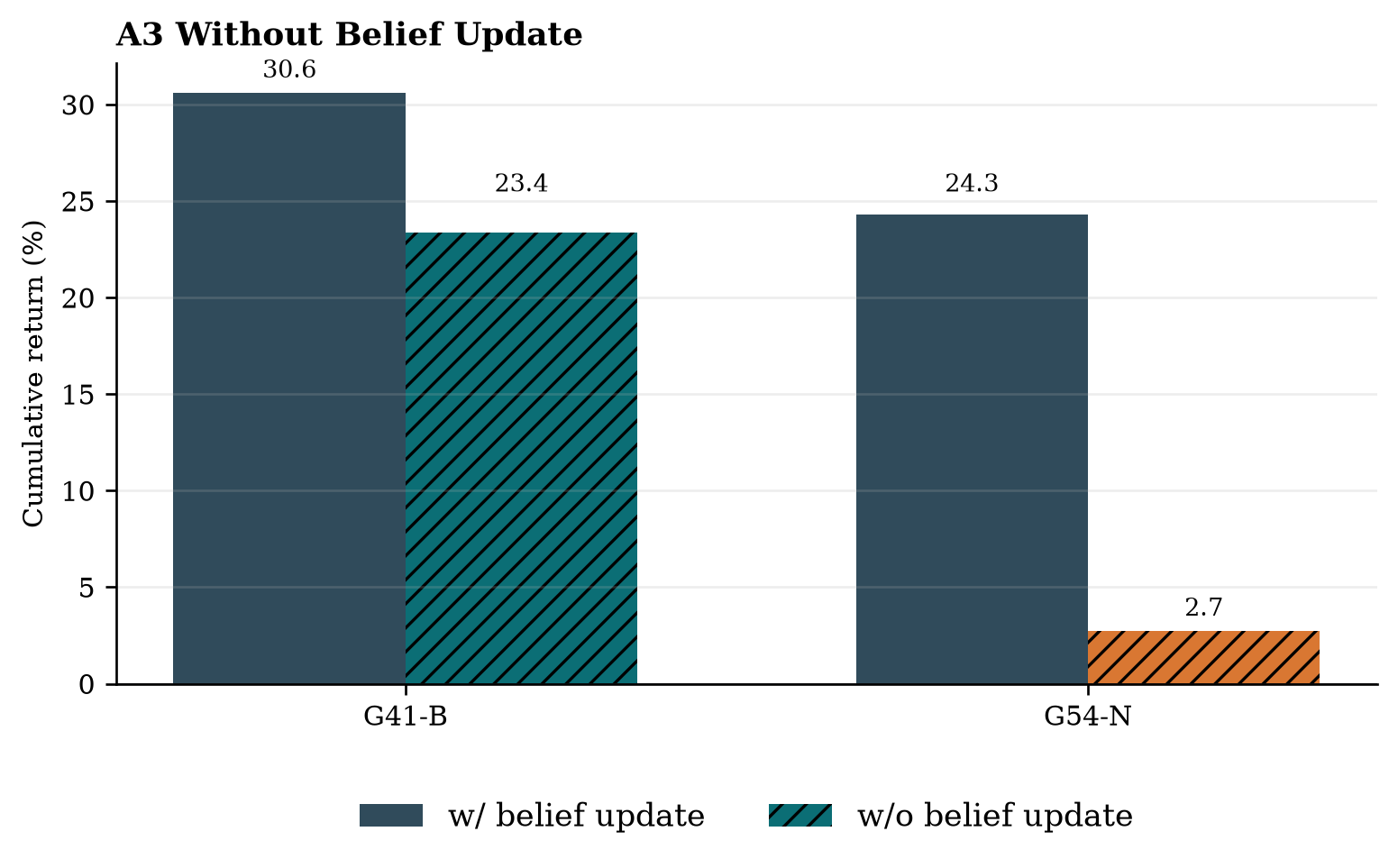}
\caption{A3 return (\%). G41-B and G54-N denote GPT-4.1-mini Beta and GPT-5.4-mini No-Beta; w/ and w/o enable/disable post-warmup belief update with shared warmup and test protocols.}
\label{fig:ablation-a3}
\par
{
\centering\small\setlength{\tabcolsep}{3.5pt}
\begin{tabular}{llrrr}
\toprule
\textbf{Setting}&\textbf{Condition}&\textbf{Ret.}&\textbf{Sh.}&\textbf{MDD}\\
\midrule
G41-B & w/ belief update &30.6&3.62&$-4.8$\\
G41-B & w/o belief update &23.4&2.51&$-10.4$\\
G54-N & w/ belief update &24.3&2.79&$-8.4$\\
G54-N & w/o belief update &2.7&0.41&$-19.2$\\
\bottomrule
\end{tabular}
\captionof{table}{A3: belief update versus warmup-only reflection memory.}
\label{tab:a3}
}
\end{figure}
Removing belief update costs GPT-5.4-mini No-Beta 21.6 return points, versus 7.2 for GPT-4.1-mini Beta, and worsens both drawdowns (Table~\ref{tab:a3}; Figure~\ref{fig:ablation-a3}). The larger loss for GPT-5.4-mini suggests greater dependence on consolidated guidance, supporting the value of validated policy updates beyond stored reflection.

\subsubsection{A4: Removing Lifecycle-Controlled Reselection}

\begin{figure}[!ht]
\centering
\includegraphics[width=\columnwidth]{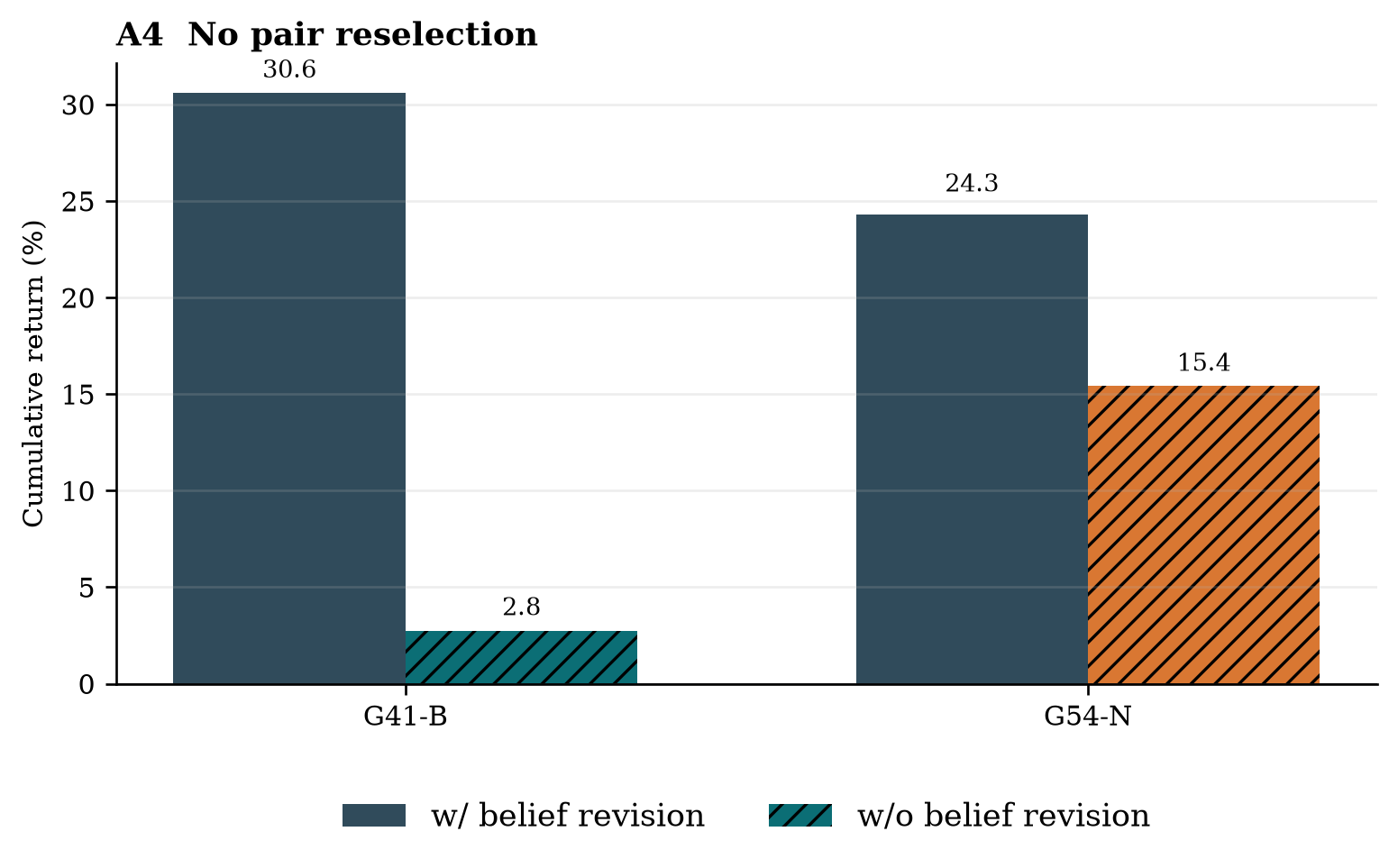}
\caption{A4 return (\%) with and without lifecycle-controlled reselection; pair trajectory is the downstream outcome.}
\label{fig:ablation-a4}
\par
{
\centering\small\setlength{\tabcolsep}{3.5pt}
\begin{tabular}{llrrr}
\toprule
\textbf{Setting}&\textbf{Condition}&\textbf{Ret.}&\textbf{Sh.}&\textbf{MDD}\\
\midrule
G41-B & w/ reselection &30.6&3.62&$-4.8$\\
G41-B & w/o reselection &2.8&0.38&$-21.8$\\
G54-N & w/ reselection &24.3&2.79&$-8.4$\\
G54-N & w/o reselection &15.4&2.02&$-7.3$\\
\bottomrule
\end{tabular}
\captionof{table}{A4: with and without lifecycle-controlled reselection.}
\label{tab:a4}
}
\end{figure}

A4 uses the main experiment's training and test windows in a same-initial-pair controlled ablation. Both variants start from the identical \method{}-selected pair and retain the same action, daily Counterfactual Reflection, and policy-memory mechanisms. The ablated variant disables only lifecycle-triggered pair replacement at validation boundaries (Equation~\ref{eq:pair-reselection}). The 23-session retry remains available only when no pair is held and cannot replace an active pair. Disabling reselection costs GPT-4.1-mini Beta 27.8 return points, versus 8.9 for GPT-5.4-mini No-Beta (Table~\ref{tab:a4}; Figure~\ref{fig:ablation-a4}), reversing their relative sensitivity in A3. This supports separate, complementary roles for policy guidance and the relationship on which it acts. \textbf{Revision has a return--risk trade-off.} Without reselection, GPT-5.4-mini's MDD improves from $-8.4\%$ to $-7.3\%$ despite lower return, whereas GPT-4.1-mini's worsens to $-21.8\%$. 

\textbf{Ablation synthesis.}
A1/A2 support persistence and regime-specific adaptability; A3/A4 isolate policy consolidation and lifecycle-controlled reselection. Their different backbone sensitivities support complementary roles for compression, validation, and hypothesis maintenance.

\subsection{Error Analysis}

Completed traces expose following three recurring error modes. All-action feedback identifies the first two; the third requires lifecycle evidence because action counterfactuals remain conditional on the held pair. 

\textbf{Excessive abstention under conflicting signals.}
The agent selects \flat{} despite a profitable directional alternative, indicating overly conservative resolution of conflicting evidence. Counterfactual Reflection records the missed payoff as corrective feedback; Appendix~\ref{app:case} (Case 1) illustrates this failure. 

\textbf{Wrong-direction continuation.}
The agent takes the losing direction while the opposite action would profit. Appendix~\ref{app:case} (Cases 2--3) traces this error and subsequent locally optimal actions, consistent with memory-conditioned correction but insufficient to establish causality. 

\textbf{Action--hypothesis attribution ambiguity.}
A corrected action can coexist with a deteriorating pair relationship, so local improvement alone cannot establish hypothesis validity. Appendix~\ref{app:case} (Case 5) shows action correction alongside lifecycle deterioration, distinguishing their temporal patterns; Appendix~\ref{app:financial-statistics} specifies the screening and break criteria used to assess relationship validity.

\section{Conclusion and Future Work}
\method{} draws on cognitive accounts in which memory is progressively reorganized, consolidated, and transformed \citep{mcclelland1995complementary,reyna1995fuzzy,winocur2011transformation}. Building upon this, our framework operationalizes agent memory as a continual process of experience compression, validation, and policy abstraction. It deliberately excludes raw observations from persistent memory, compresses action outcomes into Counterfactual Reflection records, and consolidates reward-validated experience into ticker-independent policy principles. Reward validation governs policy consolidation, while lifecycle evidence governs revision of the relationship supporting daily actions.

A3 compares belief update with warmup-only reflection memory, and A4 controls reselection, with pair trajectory as its downstream outcome. Across tested backbones and windows, these controls support complementary policy consolidation and relationship maintenance; comparison with \textsc{FinMem} shows lower token use alongside a conditional return-risk trade-off. These observations provide preliminary evidence that policy abstraction can support continual adaptation.

Future work should extend this approach toward integrated memory systems that jointly manage episodic evidence, abstraction, validation, retrieval, revision, and forgetting across multiple timescales, and evaluate them across diverse task structures and application domains to further verify this relationship.

\section*{Limitations}
Our evaluation is based on static backtests in a controlled financial domain and a limited universe of United States equities, which restricts generalization to other markets, asset classes, liquidity conditions, and decision environments. The number of independent replications remains insufficient for population-level inference, and the cross-regime study evaluates one pre-specified alternative window rather than a broad collection of market regimes. Because the evaluation treats historical market data as fixed, it does not capture the endogenous effects of deployment: trades generated by the strategy may themselves affect prices, liquidity, and the relationships between selected pairs, producing market impact and feedback that are absent from static backtests. The observation stream relies partly on model-summarized news highlights rather than raw articles, so omissions or distortions in those summaries may propagate into pair selection and action decisions. The Fixed baselines do not fully disentangle later pair selection from action adaptation, while the one-step alternatives are exact only conditional on the currently held hypothesis and do not estimate counterfactual long-run selection trajectories. Finally, results depend on the language models, prompts, aliases, and provider behavior available at evaluation time; changes to these components may limit exact reproducibility.

\section*{Ethical Considerations}
This work studies language-model agents in a financial decision setting and is intended solely for research. The reported results do not constitute investment advice, and the experimental system should not be interpreted as suitable for autonomous deployment. Real-world use could expose individuals or institutions to financial loss when model-generated reasoning, summarized news, market assumptions, or execution conditions are inaccurate or change unexpectedly. Any deployment would therefore require independent validation under realistic transaction constraints, applicable legal and compliance review, continuous monitoring, explicit risk limits, and accountable human oversight. We use TheFinAI's Herculean dataset under the CC BY 4.0 license with attribution; the terms governing its underlying yfinance/Yahoo Finance and Finnhub sources remain controlling \citep{peng2026herculean}. This paper is also released under CC BY 4.0. The authors remain responsible for the study design, interpretation, and presentation of all results.

\section*{AI Use Disclosure}
The core research idea, data selection, reference selection, model architecture, and experimental design were completed by the authors without AI assistance. For references, AI assistance was limited to checking formatting and publication metadata. Generative AI tools, including language and image-generation models, assisted with drafting and polishing the manuscript, creating and refining scientific figures, producing LaTeX tables, formatting, and layout adjustments, and writing code according to the author-designed framework and running the experiments. All AI-assisted code, experimental outputs, and manuscript material were reviewed and verified by the authors. The authors take full responsibility for the final content of this work, including all text, claims, figures, code, and reported results produced with AI assistance.

\bibliography{custom}

@inproceedings{shinn2023reflexion,
  title     = {Reflexion: Language Agents with Verbal Reinforcement Learning},
  author    = {Shinn, Noah and Cassano, Federico and Gopinath, Ashwin and Narasimhan, Karthik and Yao, Shunyu},
  booktitle = {Advances in Neural Information Processing Systems},
  volume    = {36},
  year      = {2023},
  doi       = {10.52202/075280-0377},
  url       = {https://proceedings.neurips.cc/paper_files/paper/2023/hash/1b44b878bb782e6954cd888628510e90-Abstract-Conference.html}
}

@inproceedings{madaan2023selfrefine,
  title     = {Self-Refine: Iterative Refinement with Self-Feedback},
  author    = {Madaan, Aman and Tandon, Niket and Gupta, Prakhar and Hallinan, Skyler and Gao, Luyu and Wiegreffe, Sarah and Alon, Uri and Dziri, Nouha and Prabhumoye, Shrimai and Yang, Yiming and Gupta, Shashank and Majumder, Bodhisattwa Prasad and Hermann, Katherine and Welleck, Sean and Yazdanbakhsh, Amir and Clark, Peter},
  booktitle = {Advances in Neural Information Processing Systems},
  volume    = {36},
  year      = {2023},
  doi       = {10.52202/075280-2019},
  url       = {https://proceedings.neurips.cc/paper_files/paper/2023/hash/91edff07232fb1b55a505a9e9f6c0ff3-Abstract-Conference.html}
}

@inproceedings{yao2023react,
  title     = {{ReAct}: Synergizing Reasoning and Acting in Language Models},
  author    = {Yao, Shunyu and Zhao, Jeffrey and Yu, Dian and Du, Nan and Shafran, Izhak and Narasimhan, Karthik and Cao, Yuan},
  booktitle = {International Conference on Learning Representations},
  year      = {2023},
  url       = {https://openreview.net/forum?id=WE_vluYUL-X}
}

@inproceedings{zhao2024expel,
  title     = {{ExpeL}: {LLM} Agents Are Experiential Learners},
  author    = {Zhao, Andrew and Huang, Daniel and Xu, Quentin and Lin, Matthieu and Liu, Yong-Jin and Huang, Gao},
  booktitle = {Proceedings of the AAAI Conference on Artificial Intelligence},
  volume    = {38},
  pages     = {19632--19642},
  year      = {2024},
  doi       = {10.1609/aaai.v38i17.29936},
  url       = {https://ojs.aaai.org/index.php/AAAI/article/view/29936}
}

@inproceedings{gupta2024metareflection,
  title     = {{MetaReflection}: Learning Instructions for Language Agents using Past Reflections},
  author    = {Gupta, Priyanshu and Kirtania, Shashank and Singha, Ananya and Gulwani, Sumit and Radhakrishna, Arjun and Soares, Gustavo and Shi, Sherry},
  booktitle = {Proceedings of the 2024 Conference on Empirical Methods in Natural Language Processing},
  pages     = {8369--8385},
  year      = {2024},
  publisher = {Association for Computational Linguistics},
  doi       = {10.18653/v1/2024.emnlp-main.477},
  url       = {https://aclanthology.org/2024.emnlp-main.477/}
}

@inproceedings{park2023generative,
  title     = {Generative Agents: Interactive Simulacra of Human Behavior},
  author    = {Park, Joon Sung and O'Brien, Joseph C. and Cai, Carrie Jun and Morris, Meredith Ringel and Liang, Percy and Bernstein, Michael S.},
  booktitle = {Proceedings of the 36th Annual ACM Symposium on User Interface Software and Technology},
  year      = {2023},
  publisher = {Association for Computing Machinery},
  doi       = {10.1145/3586183.3606763},
  url       = {https://doi.org/10.1145/3586183.3606763}
}

@inproceedings{maharana2024locomo,
  title     = {Evaluating Very Long-Term Conversational Memory of {LLM} Agents},
  author    = {Maharana, Adyasha and Lee, Dong-Ho and Tulyakov, Sergey and Bansal, Mohit and Barbieri, Francesco and Fang, Yuwei},
  booktitle = {Proceedings of the 62nd Annual Meeting of the Association for Computational Linguistics (Volume 1: Long Papers)},
  pages     = {13851--13870},
  year      = {2024},
  publisher = {Association for Computational Linguistics},
  doi       = {10.18653/v1/2024.acl-long.747},
  url       = {https://aclanthology.org/2024.acl-long.747/}
}

@inproceedings{tan2025membench,
  title     = {{MemBench}: Towards More Comprehensive Evaluation on the Memory of {LLM}-based Agents},
  author    = {Tan, Haoran and Zhang, Zeyu and Ma, Chen and Chen, Xu and Dai, Quanyu and Dong, Zhenhua},
  booktitle = {Findings of the Association for Computational Linguistics: ACL 2025},
  pages     = {19336--19352},
  year      = {2025},
  publisher = {Association for Computational Linguistics},
  doi       = {10.18653/v1/2025.findings-acl.989},
  url       = {https://aclanthology.org/2025.findings-acl.989/}
}

@inproceedings{tan2025reflective,
  title     = {In Prospect and Retrospect: Reflective Memory Management for Long-term Personalized Dialogue Agents},
  author    = {Tan, Zhen and Yan, Jun and Hsu, I-Hung and Han, Rujun and Wang, Zifeng and Le, Long and Song, Yiwen and Chen, Yanfei and Palangi, Hamid and Lee, George and Iyer, Anand Rajan and Chen, Tianlong and Liu, Huan and Lee, Chen-Yu and Pfister, Tomas},
  booktitle = {Proceedings of the 63rd Annual Meeting of the Association for Computational Linguistics (Volume 1: Long Papers)},
  pages     = {8416--8439},
  year      = {2025},
  publisher = {Association for Computational Linguistics},
  doi       = {10.18653/v1/2025.acl-long.413},
  url       = {https://aclanthology.org/2025.acl-long.413/}
}

@inproceedings{shen2026mem2actbench,
  title     = {{Mem2ActBench}: A Benchmark for Evaluating Long-Term Memory Utilization in Task-Oriented Autonomous Agents},
  author    = {Shen, Yiting and Li, Kun and Zhou, Wei and Hu, Songlin},
  booktitle = {Proceedings of the 64th Annual Meeting of the Association for Computational Linguistics (Volume 1: Long Papers)},
  pages     = {8173--8190},
  year      = {2026},
  publisher = {Association for Computational Linguistics},
  doi       = {10.18653/v1/2026.acl-long.370},
  url       = {https://aclanthology.org/2026.acl-long.370/}
}

@article{kaelbling1998pomdp,
  title   = {Planning and Acting in Partially Observable Stochastic Domains},
  author  = {Kaelbling, Leslie Pack and Littman, Michael L. and Cassandra, Anthony R.},
  journal = {Artificial Intelligence},
  volume  = {101},
  number  = {1--2},
  pages   = {99--134},
  year    = {1998},
  doi     = {10.1016/S0004-3702(98)00023-X},
  url     = {https://doi.org/10.1016/S0004-3702(98)00023-X}
}

@article{engle1987cointegration,
  title   = {Co-Integration and Error Correction: Representation, Estimation, and Testing},
  author  = {Engle, Robert F. and Granger, C. W. J.},
  journal = {Econometrica},
  volume  = {55},
  number  = {2},
  pages   = {251--276},
  year    = {1987},
  doi     = {10.2307/1913236},
  url     = {https://doi.org/10.2307/1913236}
}

@article{mnih2015human,
  title   = {Human-Level Control through Deep Reinforcement Learning},
  author  = {Mnih, Volodymyr and Kavukcuoglu, Koray and Silver, David and Rusu, Andrei A. and Veness, Joel and Bellemare, Marc G. and Graves, Alex and Riedmiller, Martin and Fidjeland, Andreas K. and Ostrovski, Georg and Petersen, Stig and Beattie, Charles and Sadik, Amir and Antonoglou, Ioannis and King, Helen and Kumaran, Dharshan and Wierstra, Daan and Legg, Shane and Hassabis, Demis},
  journal = {Nature},
  volume  = {518},
  pages   = {529--533},
  year    = {2015},
  doi     = {10.1038/nature14236}
}

@inproceedings{vanhasselt2016deep,
  title     = {Deep Reinforcement Learning with Double {Q}-Learning},
  author    = {van Hasselt, Hado and Guez, Arthur and Silver, David},
  booktitle = {Proceedings of the AAAI Conference on Artificial Intelligence},
  volume    = {30},
  pages     = {2094--2100},
  year      = {2016},
  doi       = {10.1609/aaai.v30i1.10295}
}

@article{gatev2006pairs,
  title   = {Pairs Trading: Performance of a Relative-Value Arbitrage Rule},
  author  = {Gatev, Evan and Goetzmann, William N. and Rouwenhorst, K. Geert},
  journal = {The Review of Financial Studies},
  volume  = {19},
  number  = {3},
  pages   = {797--827},
  year    = {2006},
  doi     = {10.1093/rfs/hhj020},
  url     = {https://doi.org/10.1093/rfs/hhj020}
}

@article{yu2025finmem,
  title   = {{FinMem}: A Performance-Enhanced {LLM} Trading Agent With Layered Memory and Character Design},
  author  = {Yu, Yangyang and Li, Haohang and Chen, Zhi and Jiang, Yuechen and Li, Yang and Suchow, Jordan W. and Zhang, Denghui and Khashanah, Khaldoun},
  journal = {IEEE Transactions on Big Data},
  volume  = {11},
  number  = {6},
  pages   = {3443--3459},
  year    = {2025},
  doi     = {10.1109/TBDATA.2025.3593370},
  url     = {https://doi.org/10.1109/TBDATA.2025.3593370}
}

@inproceedings{yu2024fincon,
  title     = {{FinCon}: A Synthesized {LLM} Multi-Agent System with Conceptual Verbal Reinforcement for Enhanced Financial Decision Making},
  author    = {Yu, Yangyang and Yao, Zhiyuan and Li, Haohang and Deng, Zhiyang and Jiang, Yuechen and Cao, Yupeng and Chen, Zhi and Suchow, Jordan W. and Cui, Zhenyu and Liu, Rong and Xu, Zhaozhuo and Zhang, Denghui and Subbalakshmi, Koduvayur and Xiong, Guojun and He, Yueru and Huang, Jimin and Li, Dong and Xie, Qianqian},
  booktitle = {Advances in Neural Information Processing Systems},
  volume    = {37},
  year      = {2024},
  doi       = {10.52202/079017-4354},
  url       = {https://proceedings.neurips.cc/paper_files/paper/2024/hash/f7ae4fe91d96f50abc2211f09b6a7e49-Abstract-Conference.html}
}

@inproceedings{zhang2024finagent,
  title     = {A Multimodal Foundation Agent for Financial Trading: Tool-Augmented, Diversified, and Generalist},
  author    = {Zhang, Wentao and Zhao, Lingxuan and Xia, Haochong and Sun, Shuo and Sun, Jiaze and Qin, Molei and Li, Xinyi and Zhao, Yuqing and Zhao, Yilei and Cai, Xinyu and Zheng, Longtao and Wang, Xinrun and An, Bo},
  booktitle = {Proceedings of the 30th ACM SIGKDD Conference on Knowledge Discovery and Data Mining},
  pages     = {4314--4325},
  year      = {2024},
  publisher = {Association for Computing Machinery},
  doi       = {10.1145/3637528.3671801},
  url       = {https://doi.org/10.1145/3637528.3671801}
}

@inproceedings{xiao2024tradingagents,
  title   = {{TradingAgents}: Multi-Agents {LLM} Financial Trading Framework},
  author  = {Xiao, Yijia and Sun, Edward and Luo, Di and Wang, Wei},
  booktitle = {Multi-Agent AI in the Real World Workshop at AAAI 2025},
  year    = {2025},
  url     = {https://openreview.net/forum?id=4QPrXwMQt1}
}

@inproceedings{wu2024streambench,
  title     = {{StreamBench}: Towards Benchmarking Continuous Improvement of Language Agents},
  author    = {Wu, Cheng-Kuang and Tam, Zhi Rui and Lin, Chieh-Yen and Chen, Yun-Nung and Lee, Hung-yi},
  booktitle = {Advances in Neural Information Processing Systems},
  volume    = {37},
  year      = {2024},
  doi       = {10.52202/079017-3398},
  url       = {https://proceedings.neurips.cc/paper_files/paper/2024/hash/c189915371c4474fe9789be3728113fc-Abstract-Datasets_and_Benchmarks_Track.html}
}

@inproceedings{wu2025longmemeval,
  title     = {{LongMemEval}: Benchmarking Chat Assistants on Long-Term Interactive Memory},
  author    = {Wu, Di and Wang, Hongwei and Yu, Wenhao and Zhang, Yuwei and Chang, Kai-Wei and Yu, Dong},
  booktitle = {International Conference on Learning Representations},
  year      = {2025},
  url       = {https://proceedings.iclr.cc/paper_files/paper/2025/hash/d813d324dbf0598bbdc9c8e79740ed01-Abstract-Conference.html}
}

@inproceedings{qin2026implicit,
  title     = {{ImplicitMemBench}: Measuring Unconscious Behavioral Adaptation in Large Language Models},
  author    = {Qin, Chonghan and Feng, Xiachong and Ma, Weitao and Feng, Xiaocheng and Kong, Lingpeng},
  booktitle = {Proceedings of the 64th Annual Meeting of the Association for Computational Linguistics (Volume 1: Long Papers)},
  pages     = {28232--28261},
  year      = {2026},
  publisher = {Association for Computational Linguistics},
  doi       = {10.18653/v1/2026.acl-long.1301},
  url       = {https://aclanthology.org/2026.acl-long.1301/}
}

@inproceedings{liu2024agentbench,
  title     = {{AgentBench}: Evaluating {LLM}s as Agents},
  author    = {Liu, Xiao and Yu, Hao and Zhang, Hanchen and Xu, Yifan and Lei, Xuanyu and Lai, Hanyu and Gu, Yu and Ding, Hangliang and Men, Kaiwen and Yang, Kejuan and Zhang, Shudan and Deng, Xiang and Zeng, Aohan and Du, Zhengxiao and Zhang, Chenhui and Shen, Sheng and Zhang, Tianjun and Su, Yu and Sun, Huan and Huang, Minlie and Dong, Yuxiao and Tang, Jie},
  booktitle = {International Conference on Learning Representations},
  year      = {2024},
  url       = {https://openreview.net/forum?id=zAdUB0aCTQ}
}

@inproceedings{ma2024agentboard,
  title     = {{AgentBoard}: An Analytical Evaluation Board of Multi-turn {LLM} Agents},
  author    = {Ma, Chang and Zhang, Junlei and Zhu, Zhihao and Yang, Cheng and Yang, Yujiu and Jin, Yaohui and Lan, Zhenzhong and Kong, Lingpeng and He, Junxian},
  booktitle = {Advances in Neural Information Processing Systems},
  volume    = {37},
  year      = {2024},
  doi       = {10.52202/079017-2365},
  url       = {https://proceedings.neurips.cc/paper_files/paper/2024/hash/877b40688e330a0e2a3fc24084208dfa-Abstract-Datasets_and_Benchmarks_Track.html}
}

@inproceedings{mialon2024gaia,
  title     = {{GAIA}: A Benchmark for General {AI} Assistants},
  author    = {Mialon, Gr{\'e}goire and Fourrier, Cl{\'e}mentine and Swift, Craig and Wolf, Thomas and LeCun, Yann and Scialom, Thomas},
  booktitle = {International Conference on Learning Representations},
  year      = {2024},
  url       = {https://openreview.net/forum?id=fibxvahvs3}
}

@inproceedings{qin2024toolllm,
  title     = {{ToolLLM}: Facilitating Large Language Models to Master 16000+ Real-world {API}s},
  author    = {Qin, Yujia and Liang, Shihao and Ye, Yining and Zhu, Kunlun and Yan, Lan and Lu, Yaxi and Lin, Yankai and Cong, Xin and Tang, Xiangru and Qian, Bill and Zhao, Sihan and Hong, Lauren and Tian, Runchu and Xie, Ruobing and Zhou, Jie and Gerstein, Mark and Li, Dahai and Liu, Zhiyuan and Sun, Maosong},
  booktitle = {International Conference on Learning Representations},
  year      = {2024},
  url       = {https://openreview.net/forum?id=dHng2O0Jjr}
}

@article{dickey1979unitroot,
  title   = {Distribution of the Estimators for Autoregressive Time Series with a Unit Root},
  author  = {Dickey, David A. and Fuller, Wayne A.},
  journal = {Journal of the American Statistical Association},
  volume  = {74},
  number  = {366a},
  pages   = {427--431},
  year    = {1979},
  doi     = {10.1080/01621459.1979.10482531},
  url     = {https://doi.org/10.1080/01621459.1979.10482531}
}

@article{said1984unitroots,
  title   = {Testing for Unit Roots in Autoregressive-Moving Average Models of Unknown Order},
  author  = {Said, Said E. and Dickey, David A.},
  journal = {Biometrika},
  volume  = {71},
  number  = {3},
  pages   = {599--607},
  year    = {1984},
  doi     = {10.1093/biomet/71.3.599},
  url     = {https://doi.org/10.1093/biomet/71.3.599}
}

@article{uhlenbeck1930brownian,
  title   = {On the Theory of the Brownian Motion},
  author  = {Uhlenbeck, George E. and Ornstein, Leonard S.},
  journal = {Physical Review},
  volume  = {36},
  number  = {5},
  pages   = {823--841},
  year    = {1930},
  doi     = {10.1103/PhysRev.36.823},
  url     = {https://doi.org/10.1103/PhysRev.36.823}
}

@article{elliott2005pairs,
  title   = {Pairs Trading},
  author  = {Elliott, Robert J. and van der Hoek, John and Malcolm, William P.},
  journal = {Quantitative Finance},
  volume  = {5},
  number  = {3},
  pages   = {271--276},
  year    = {2005},
  doi     = {10.1080/14697680500149370},
  url     = {https://doi.org/10.1080/14697680500149370}
}

@inproceedings{zhong2024memorybank,
  title     = {{MemoryBank}: Enhancing Large Language Models with Long-Term Memory},
  author    = {Zhong, Wanjun and Guo, Lianghong and Gao, Qiqi and Ye, He and Wang, Yanlin},
  booktitle = {Proceedings of the AAAI Conference on Artificial Intelligence},
  volume    = {38},
  pages     = {19724--19731},
  year      = {2024},
  doi       = {10.1609/aaai.v38i17.29946},
  url       = {https://ojs.aaai.org/index.php/AAAI/article/view/29946}
}

@article{packer2023memgpt,
  title   = {{MemGPT}: Towards {LLM}s as Operating Systems},
  author  = {Packer, Charles and Wooders, Sarah and Lin, Kevin and Fang, Vivian and Patil, Shishir G. and Stoica, Ion and Gonzalez, Joseph E.},
  journal = {arXiv preprint arXiv:2310.08560},
  year    = {2023},
  doi     = {10.48550/arXiv.2310.08560},
  url     = {https://arxiv.org/abs/2310.08560}
}

@inproceedings{wang2024emgrag,
  title     = {Crafting Personalized Agents through Retrieval-Augmented Generation on Editable Memory Graphs},
  author    = {Wang, Zheng and Li, Zhongyang and Jiang, Zeren and Tu, Dandan and Shi, Wei},
  booktitle = {Proceedings of the 2024 Conference on Empirical Methods in Natural Language Processing},
  pages     = {4891--4906},
  year      = {2024},
  publisher = {Association for Computational Linguistics},
  doi       = {10.18653/v1/2024.emnlp-main.281},
  url       = {https://aclanthology.org/2024.emnlp-main.281/}
}

@inproceedings{kang2025memoryos,
  title     = {Memory {OS} of {AI} Agent},
  author    = {Kang, Jiazheng and Ji, Mingming and Zhao, Zhe and Bai, Ting},
  booktitle = {Proceedings of the 2025 Conference on Empirical Methods in Natural Language Processing},
  pages     = {25961--25970},
  year      = {2025},
  publisher = {Association for Computational Linguistics},
  doi       = {10.18653/v1/2025.emnlp-main.1318},
  url       = {https://aclanthology.org/2025.emnlp-main.1318/}
}

@inproceedings{xu2025amem,
  title     = {{A-Mem}: Agentic Memory for {LLM} Agents},
  author    = {Xu, Wujiang and Liang, Zujie and Mei, Kai and Gao, Hang and Tan, Juntao and Zhang, Yongfeng},
  booktitle = {Advances in Neural Information Processing Systems},
  volume    = {38},
  year      = {2025},
  doi       = {10.52202/085713-0593},
  url       = {https://proceedings.neurips.cc/paper_files/paper/2025/hash/19909c36f51abc4856b4560aff3d36d6-Abstract-Conference.html}
}

@inproceedings{latimer2026hindsight,
  title     = {Hindsight: Structured Agent Memory that Retains, Recalls, and Reflects},
  author    = {Latimer, Christopher and Boschi, Nicol{\`o} and Neeser, Andrew and Bartholomew, Chris and Srivastava, Gaurav and Wang, Xuan and Ramakrishnan, Naren},
  booktitle = {Proceedings of the 64th Annual Meeting of the Association for Computational Linguistics (Volume 3: System Demonstrations)},
  pages     = {275--285},
  year      = {2026},
  publisher = {Association for Computational Linguistics},
  doi       = {10.18653/v1/2026.acl-demo.27},
  url       = {https://aclanthology.org/2026.acl-demo.27/}
}

@inproceedings{luo2026memorysurvey,
  title     = {From Storage to Experience: A Survey on the Evolution of {LLM} Agent Memory Mechanisms},
  author    = {Luo, Jinghao and Tian, Yuchen and Cao, Chuxue and Luo, Ziyang and Lin, Hongzhan and Li, Kaixin and Kong, Chuyi and Yang, Ruichao and Ma, Jing},
  booktitle = {Findings of the Association for Computational Linguistics: ACL 2026},
  pages     = {41622--41652},
  year      = {2026},
  publisher = {Association for Computational Linguistics},
  doi       = {10.18653/v1/2026.findings-acl.2069},
  url       = {https://aclanthology.org/2026.findings-acl.2069/}
}

@article{wang2024voyager,
  title   = {Voyager: An Open-Ended Embodied Agent with Large Language Models},
  author  = {Wang, Guanzhi and Xie, Yuqi and Jiang, Yunfan and Mandlekar, Ajay and Xiao, Chaowei and Zhu, Yuke and Fan, Linxi and Anandkumar, Anima},
  journal = {Transactions on Machine Learning Research},
  year    = {2024},
  url     = {https://openreview.net/forum?id=ehfRiF0R3a}
}

@inproceedings{wang2025awm,
  title     = {Agent Workflow Memory},
  author    = {Wang, Zora Zhiruo and Mao, Jiayuan and Fried, Daniel and Neubig, Graham},
  booktitle = {Proceedings of the 42nd International Conference on Machine Learning},
  pages     = {63897--63911},
  volume    = {267},
  series    = {Proceedings of Machine Learning Research},
  publisher = {PMLR},
  year      = {2025},
  url       = {https://proceedings.mlr.press/v267/wang25bx.html}
}

@inproceedings{ma2026finemem,
  title     = {{Fine-Mem}: Fine-Grained Feedback Alignment for Long-Horizon Memory Management},
  author    = {Ma, Weitao and Feng, Xiaocheng and Huang, Lei and Feng, Xiachong and Ma, Zhanyu and Xu, Jun and Gao, Jiuchong and Hao, Jinghua and He, Renqing and Qin, Bing},
  booktitle = {Proceedings of the 64th Annual Meeting of the Association for Computational Linguistics (Volume 1: Long Papers)},
  pages     = {19666--19684},
  year      = {2026},
  publisher = {Association for Computational Linguistics},
  doi       = {10.18653/v1/2026.acl-long.900},
  url       = {https://aclanthology.org/2026.acl-long.900/}
}

@article{sutton1991dyna,
  title   = {Dyna, an Integrated Architecture for Learning, Planning, and Reacting},
  author  = {Sutton, Richard S.},
  journal = {ACM SIGART Bulletin},
  volume  = {2},
  number  = {4},
  pages   = {160--163},
  year    = {1991},
  doi     = {10.1145/122344.122377},
  url     = {https://doi.org/10.1145/122344.122377}
}

@inproceedings{andrychowicz2017her,
  title     = {Hindsight Experience Replay},
  author    = {Andrychowicz, Marcin and Wolski, Filip and Ray, Alex and Schneider, Jonas and Fong, Rachel and Welinder, Peter and McGrew, Bob and Tobin, Josh and Abbeel, Pieter and Zaremba, Wojciech},
  booktitle = {Advances in Neural Information Processing Systems},
  volume    = {30},
  year      = {2017},
  url       = {https://proceedings.neurips.cc/paper/2017/hash/453fadbd8a1a3af50a9df4df899537b5-Abstract.html}
}

@book{sutton2018rl,
  title     = {Reinforcement Learning: An Introduction},
  author    = {Sutton, Richard S. and Barto, Andrew G.},
  edition   = {2},
  publisher = {MIT Press},
  year      = {2018},
  url       = {https://mitpress.mit.edu/9780262039246/reinforcement-learning/}
}

@inproceedings{zhou2024webarena,
  title     = {{WebArena}: A Realistic Web Environment for Building Autonomous Agents},
  author    = {Zhou, Shuyan and Xu, Frank F. and Zhu, Hao and Zhou, Xuhui and Lo, Robert and Sridhar, Abishek and Cheng, Xianyi and Ou, Tianyue and Bisk, Yonatan and Fried, Daniel and Alon, Uri and Neubig, Graham},
  booktitle = {International Conference on Learning Representations},
  year      = {2024},
  url       = {https://openreview.net/forum?id=oKn9c6ytLx}
}

@inproceedings{deng2023mind2web,
  title     = {{Mind2Web}: Towards a Generalist Agent for the Web},
  author    = {Deng, Xiang and Gu, Yu and Zheng, Boyuan and Chen, Shijie and Stevens, Sam and Wang, Boshi and Sun, Huan and Su, Yu},
  booktitle = {Advances in Neural Information Processing Systems},
  volume    = {36},
  pages     = {28091--28114},
  year      = {2023},
  doi       = {10.52202/075280-1220},
  url       = {https://proceedings.neurips.cc/paper_files/paper/2023/hash/5950bf290a1570ea401bf98882128160-Abstract-Datasets_and_Benchmarks.html}
}

@inproceedings{xie2024osworld,
  title     = {{OSWorld}: Benchmarking Multimodal Agents for Open-Ended Tasks in Real Computer Environments},
  author    = {Xie, Tianbao and Zhang, Danyang and Chen, Jixuan and Li, Xiaochuan and Zhao, Siheng and Cao, Ruisheng and Hua, Toh Jing and Cheng, Zhoujun and Shin, Dongchan and Lei, Fangyu and Liu, Yitao and Xu, Yiheng and Zhou, Shuyan and Savarese, Silvio and Xiong, Caiming and Zhong, Victor and Yu, Tao},
  booktitle = {Advances in Neural Information Processing Systems},
  volume    = {37},
  pages     = {52040--52094},
  year      = {2024},
  doi       = {10.52202/079017-1650},
  url       = {https://doi.org/10.52202/079017-1650}
}

@inproceedings{xie2023pixiu,
  title     = {{PIXIU}: A Comprehensive Benchmark, Instruction Dataset and Large Language Model for Finance},
  author    = {Xie, Qianqian and Han, Weiguang and Zhang, Xiao and Lai, Yanzhao and Peng, Min and Lopez-Lira, Alejandro and Huang, Jimin},
  booktitle = {Advances in Neural Information Processing Systems},
  volume    = {36},
  pages     = {33469--33484},
  year      = {2023},
  doi       = {10.52202/075280-1454},
  url       = {https://doi.org/10.52202/075280-1454}
}

@inproceedings{xie2024finben,
  title     = {{FinBen}: A Holistic Financial Benchmark for Large Language Models},
  author    = {Xie, Qianqian and Han, Weiguang and Chen, Zhengyu and Xiang, Ruoyu and Zhang, Xiao and He, Yueru and Xiao, Mengxi and Li, Dong and Dai, Yongfu and Feng, Duanyu and Xu, Yijing and Kang, Haoqiang and Kuang, Ziyan and Yuan, Chenhan and Yang, Kailai and Luo, Zheheng and Zhang, Tianlin and Liu, Zhiwei and Xiong, Guojun and Deng, Zhiyang and Jiang, Yuechen and Yao, Zhiyuan and Li, Haohang and Yu, Yangyang and Hu, Gang and Huang, Jiajia and Liu, Xiao-Yang and Lopez-Lira, Alejandro and Wang, Benyou and Lai, Yanzhao and Wang, Hao and Peng, Min and Ananiadou, Sophia and Huang, Jimin},
  booktitle = {Advances in Neural Information Processing Systems},
  volume    = {37},
  pages     = {95716--95743},
  year      = {2024},
  doi       = {10.52202/079017-3033},
  url       = {https://doi.org/10.52202/079017-3033}
}

@article{avellaneda2010statistical,
  title   = {Statistical Arbitrage in the {US} Equities Market},
  author  = {Avellaneda, Marco and Lee, Jeong-Hyun},
  journal = {Quantitative Finance},
  volume  = {10},
  number  = {7},
  pages   = {761--782},
  year    = {2010},
  doi     = {10.1080/14697680903124632},
  url     = {https://doi.org/10.1080/14697680903124632}
}

@misc{peng2026herculean,
  title        = {{Herculean}: An Offline Multi-Skill Evaluation Benchmark for {LLM} Agents in Finance},
  author       = {Peng, Xueqing},
  year         = {2026},
  howpublished = {Hugging Face dataset},
  url          = {https://huggingface.co/datasets/TheFinAI/Herculean},
  note         = {Financial agentic benchmark with offline trading, hedging, report, and XBRL auditing tasks; CC BY 4.0}
}

@misc{openai2025gpt41,
  title        = {Introducing {GPT-4.1} in the {API}},
  author       = {{OpenAI}},
  year         = {2025},
  month        = apr,
  howpublished = {Official model release report},
  url          = {https://openai.com/index/gpt-4-1/}
}

@misc{openai2026gpt54small,
  title        = {Introducing {GPT-5.4} Mini and Nano},
  author       = {{OpenAI}},
  year         = {2026},
  month        = mar,
  howpublished = {Official model release report},
  url          = {https://openai.com/index/introducing-gpt-5-4-mini-and-nano/}
}

@misc{openai2026gpt54card,
  title        = {{GPT-5.4} Thinking System Card},
  author       = {{OpenAI}},
  year         = {2026},
  month        = mar,
  howpublished = {Technical report},
  url          = {https://deploymentsafety.openai.com/gpt-5-4-thinking/gpt-5-4-thinking.pdf}
}

@misc{openai2025o4mini,
  title        = {{OpenAI o3} and {o4-mini} System Card},
  author       = {{OpenAI}},
  year         = {2025},
  month        = apr,
  howpublished = {Technical report},
  url          = {https://cdn.openai.com/pdf/2221c875-02dc-4789-800b-e7758f3722c1/o3-and-o4-mini-system-card.pdf}
}

@article{deepseek2026v4,
  title   = {{DeepSeek-V4.1-Flash}: Pushing the Limits of {KV} Cache Compression},
  author  = {{DeepSeek-AI}},
  journal = {arXiv preprint arXiv:2609.19969},
  year    = {2026},
  doi     = {10.48550/arXiv.2609.19969},
  url     = {https://arxiv.org/abs/2609.19969},
  note    = {Technical report; the full author list exceeds 65 authors}
}

@article{sharpe1994ratio,
  title   = {The Sharpe Ratio},
  author  = {Sharpe, William F.},
  journal = {The Journal of Portfolio Management},
  volume  = {21},
  number  = {1},
  pages   = {49--58},
  year    = {1994},
  doi     = {10.3905/jpm.1994.409501},
  url     = {https://doi.org/10.3905/jpm.1994.409501}
}

@article{lo2002statistics,
  author = {Andrew W. Lo},
  title = {The Statistics of {Sharpe} Ratios},
  journal = {Financial Analysts Journal},
  year = {2002},
  volume = {58},
  number = {4},
  pages = {36--52},
  doi = {10.2469/faj.v58.n4.2453},
  url = {https://doi.org/10.2469/faj.v58.n4.2453}
}

@article{chekhlov2005drawdown,
  title   = {Drawdown Measure in Portfolio Optimization},
  author  = {Chekhlov, Alexei and Uryasev, Stanislav and Zabarankin, Michael},
  journal = {International Journal of Theoretical and Applied Finance},
  volume  = {8},
  number  = {1},
  pages   = {13--58},
  year    = {2005},
  doi     = {10.1142/S0219024905002767},
  url     = {https://doi.org/10.1142/S0219024905002767}
}

@article{demiguel2009naive,
  title   = {Optimal versus Naive Diversification: How Inefficient Is the {1/N} Portfolio Strategy?},
  author  = {DeMiguel, Victor and Garlappi, Lorenzo and Uppal, Raman},
  journal = {The Review of Financial Studies},
  volume  = {22},
  number  = {5},
  pages   = {1915--1953},
  year    = {2009},
  doi     = {10.1093/rfs/hhm075},
  url     = {https://doi.org/10.1093/rfs/hhm075}
}

@article{mcclelland1995complementary,
  author  = {James L. McClelland and Bruce L. McNaughton and Randall C. O'Reilly},
  title   = {Why There Are Complementary Learning Systems in the Hippocampus and Neocortex: Insights from the Successes and Failures of Connectionist Models of Learning and Memory},
  journal = {Psychological Review},
  year    = {1995},
  volume  = {102},
  number  = {3},
  pages   = {419--457},
  doi     = {10.1037/0033-295X.102.3.419},
  url     = {https://doi.org/10.1037/0033-295X.102.3.419}
}

@article{reyna1995fuzzy,
  author  = {Valerie F. Reyna and Charles J. Brainerd},
  title   = {Fuzzy-Trace Theory: An Interim Synthesis},
  journal = {Learning and Individual Differences},
  year    = {1995},
  volume  = {7},
  number  = {1},
  pages   = {1--75},
  doi     = {10.1016/1041-6080(95)90031-4},
  url     = {https://doi.org/10.1016/1041-6080(95)90031-4}
}

@article{winocur2011transformation,
  author  = {Gordon Winocur and Morris Moscovitch},
  title   = {Memory Transformation and Systems Consolidation},
  journal = {Journal of the International Neuropsychological Society},
  year    = {2011},
  volume  = {17},
  number  = {5},
  pages   = {766--780},
  doi     = {10.1017/S1355617711000683},
  url     = {https://doi.org/10.1017/S1355617711000683}
}

@inproceedings{yu2026agentic,
  title = {Agentic Memory: Learning Unified Long-Term and Short-Term Memory Management for Large Language Model Agents},
  author = {Yu, Yi and Yao, Liuyi and Xie, Yuexiang and Tan, Qingquan and Feng, Jiaqi and Li, Yaliang and Wu, Libing},
  booktitle = {Proceedings of the 64th Annual Meeting of the Association for Computational Linguistics (Volume 1: Long Papers)},
  year = {2026},
  month = jul,
  address = {San Diego, California, United States},
  publisher = {Association for Computational Linguistics},
  pages = {21457--21483},
  doi = {10.18653/v1/2026.acl-long.981},
  url = {https://aclanthology.org/2026.acl-long.981/}
}

@misc{zhu2026knowingdoing,
  title = {From Knowing to Doing: A Memory-Controlled Benchmark for {LLM} Trading Agents on Stock Markets},
  author = {Zhu, Taojie and Zhao, Wentao and Sun, Rui and Luan, Beidi and Lu, Jiacheng and Wang, Sinuo and Li, Jing and Jiang, Daxin and He, Yonghong and Bai, Zuo},
  year = {2026},
  eprint = {2605.28359},
  archivePrefix = {arXiv},
  primaryClass = {cs.AI},
  doi = {10.48550/arXiv.2605.28359},
  url = {https://arxiv.org/abs/2605.28359}
}

@misc{google2026gemini31flashlite,
  author = {{Google DeepMind}},
  title = {{Gemini 3.1 Flash-Lite}},
  year = {2026},
  month = mar,
  howpublished = {Official model card},
  url = {https://deepmind.google/models/model-cards/gemini-3-1-flash-lite/}
}

@techreport{geminiteam2023gemini,
  author = {{Gemini Team}},
  title = {{Gemini}: A Family of Highly Capable Multimodal Models},
  institution = {Google DeepMind},
  year = {2023},
  number = {arXiv:2312.11805},
  url = {https://arxiv.org/abs/2312.11805}
}

@misc{google2026gemini38flash,
  author = {{Google DeepMind}},
  title = {{Gemini 3.8 Flash}},
  year = {2026},
  month = sep,
  howpublished = {Official model card},
  url = {https://deepmind.google/models/model-cards/gemini-3-8-flash/}
}

@misc{meta2024llama31,
  author = {{Meta}},
  title = {{Llama 3.1 Instruct Model Card}},
  year = {2024},
  howpublished = {Official model card},
  url = {https://huggingface.co/meta-llama/Llama-3.1-8B-Instruct}
}

@misc{qwen2025qwen34b,
  author = {{Qwen Team}},
  title = {{Qwen3-4B}},
  year = {2025},
  howpublished = {Official model card},
  url = {https://huggingface.co/Qwen/Qwen3-4B}
}

@misc{qwen2025qwen332b,
  author = {{Qwen Team}},
  title = {{Qwen3-32B}},
  year = {2025},
  howpublished = {Official model card},
  url = {https://huggingface.co/Qwen/Qwen3-32B}
}

\appendix
\raggedbottom
\section{Complete Experiment Prompts}
\label{app:prompts}
This section lists the current language-model system prompts for the primary experiment and A1--A4. In the prompts, PnL, ADF, M\&A, and JSON denote profit and loss, augmented Dickey--Fuller, mergers and acquisitions, and JavaScript Object Notation.

\subsection{Pair-Selection System Prompt}
This prompt selects one pair from the causal shortlist; action selection is a separate call.
\begin{Verbatim}[fontsize=\normalsize,breaklines=true,breakanywhere=true]
PAIR SELECTION TASK

You are selecting a pair for a statistical pair-trading strategy.

Pair trading seeks two assets whose prices have a stable long-run relationship. A good pair should have:

1. strong historical cointegration,
2. a reasonable mean-reversion speed,
3. interpretable economic or business relationship,
4. sufficient spread movement to trade,
5. no obvious structural breakdown.

The candidate pairs below were generated from a
causal 10-year statistical screen. Your task is
not to compute cointegration. Your task is to
choose the most attractive pair from the
candidate pool.

When selecting a pair, consider:

- cointegration p-value,
- half-life,
- beta / hedge ratio,
- long-horizon z-score,
- trading-window z-score,
- current lifecycle state,
- pair memory success/failure history,
- business relationship between the two companies,
- whether their revenue models or business exposures make the pair economically meaningful.

Company context is provided for each stock, limited to its main business; undated financial results are excluded.

Choose ONE pair only.

Return JSON:

{
  "selected_pair": "AAA/BBB",
  "pair_selection_reason": "..."
}
\end{Verbatim}

\subsection{Daily-Direction System Prompt}
This base prompt governs daily actions on the active pair.
\begin{Verbatim}[fontsize=\normalsize,breaklines=true,breakanywhere=true]
PAIR TRADING DIRECTION DISCOVERY TASK

Pair Selection has already been completed. Do not choose a new pair.

You are now deciding how to trade the active pair today.

In pair trading, direction is about relative performance between Asset A and Asset B.

Actions:

LONG_SHORT:
- long Asset A
- short Asset B
- use this when Asset A is expected to outperform Asset B, or when the spread suggests Asset A is cheap relative to Asset B and likely to revert upward.

SHORT_LONG:
- short Asset A
- long Asset B
- use this when Asset B is expected to outperform Asset A, or when the spread suggests Asset A is rich relative to Asset B and likely to revert downward.

FLAT:
- hold cash
- use only when the evidence is weak, contradictory, or lifecycle constraints make trading unsafe.

Your objective is to maximize risk-adjusted return, not to minimize trading activity.

You must consider:

1. current spread state,
2. long-horizon z-score,
3. trading-window z-score,
4. today's prices,
5. today's price deltas: P_T - P_{T-1},
6. latest available news,
7. relative news between Asset A and Asset B,
8. company business context,
9. economic relationship between the companies,
10. past 3 days Direction Memory,
11. lifecycle state.

Do not analyze Asset A and Asset B independently. Compare them.

Ask:

- Which asset is likely to outperform the other today?
- Does the spread suggest mean reversion or continued divergence?
- Does today's price delta confirm or contradict the spread signal?
- Did recent memory show that similar decisions were right or wrong?
- Did the counterfactual actions outperform the chosen action in recent days?

Lifecycle state is contextual information only;
it is never an action constraint. WARNING means
the pair relationship is weakening but not
confirmed broken; it must not force FLAT or
prevent trading. BREAK means the pair has failed
pair-quality validation and will be replaced at
the next pair-selection opportunity; it warrants
extra caution, but it does NOT force FLAT, does
NOT prohibit trading, and does NOT replace your
decision with a fixed action. Under any
lifecycle state, choose one of LONG_SHORT /
SHORT_LONG / FLAT using the same process: spread
state, price movement, news, and recent memory.

FLAT is valid, but repeated FLAT decisions should not become the default policy. If there is a clear relative edge, choose LONG_SHORT or SHORT_LONG.

Return JSON:

{
  "action": "LONG_SHORT" | "SHORT_LONG" | "FLAT",
  "direction_reason": "..."
}
\end{Verbatim}

\subsection{Training-Reflection System Prompt}
This prompt runs after an eligible training round to propose transferable principles.
\begin{Verbatim}[fontsize=\normalsize,breaklines=true,breakanywhere=true]
You are a trading coach. From the round's DAILY decisions
with COUNTERFACTUAL PnL (what LONG_SHORT/SHORT_LONG/FLAT
would each have earned) write a SHORT reflection (<=6 lines)
and 1-3 TRANSFERABLE decision PRINCIPLES that improve future
direction decisions. NEVER mention a specific ticker, pair,
date, step, or a specific LONG_SHORT/SHORT_LONG
recommendation. Learn from: how often the chosen action
matched the counterfactual best action, repeated wrong-side
decisions, whether one direction was systematically better,
whether FLAT was over- or under-used, whether the daily
price delta contradicted spread reversion, and whether news
supported continuation or reversion. End with a line
'Principles:' followed by the principles, one per line.
\end{Verbatim}

\subsection{Relationship-Break System Prompt}
This prompt supplies semantic news evidence at a scheduled lifecycle validation boundary; the code applies the final state threshold and replacement rule.
\begin{Verbatim}[fontsize=\normalsize,breaklines=true,breakanywhere=true]
You assess the PAST-7-DAY NEWS for an ACTIVE statistical-arbitrage PAIR (leg A, leg B) and judge whether the news threatens the mean-reverting RELATIONSHIP.
Evaluate the SEMANTIC RELATIONSHIP FIRST, then the FINANCIAL-MARKET SENTIMENT.
Provide CONFIRMING evidence only -- the final BREAK is enforced by code rules.

A. SEMANTIC RELATIONSHIP (first):
   - Does the news affect only ONE leg (asymmetric business risk)?
   - Does it weaken the economic linkage between the two stocks?
   - Does it increase structural divergence or reduce mean-reversion plausibility?

B. FINANCIAL SENTIMENT (second; financial-market sentiment, NOT emotional tone):
   Positive: earnings beat, guidance raise, analyst upgrade, strong demand,
     market-share gain, strategic advantage, short sellers turning bullish.
   Negative: earnings miss, guidance cut, analyst downgrade, demand weakness,
     regulatory pressure, accounting concern, product failure, lawsuit,
     management instability, short sellers turning more bearish.

Rules (3-level HEALTHY / WARNING / BREAK -- no LOW/MEDIUM/HIGH/CRITICAL):
   - stable semantic relationship + neutral or mixed sentiment -> HEALTHY;
   - mild pair-unfavorable semantic news -> WARNING;
   - BOTH legs same strong direction (both strong_positive OR both strong_negative)
     -> at least WARNING (same strong direction reduces market-neutral usefulness);
   - one strong_positive and the other neutral/negative -> at least WARNING;
   - one strong_negative and the other neutral/positive -> at least WARNING;
   - a direct relationship-breaking event, or structural divergence that
     invalidates the pair logic -> BREAK.

Return ONLY JSON: {"semantic_relationship_signal": "stable|weakening|broken_candidate", "leg_a_financial_sentiment": "strong_positive|positive|neutral|negative|strong_negative", "leg_b_financial_sentiment": "strong_positive|positive|neutral|negative|strong_negative", "sentiment_symmetry": "same_strong_direction|asymmetric|mixed_or_neutral", "news_break_level": "HEALTHY|WARNING|BREAK", "is_relationship_break_signal": true|false, "reason": "brief explanation"}
\end{Verbatim}

\subsection{Critical-News Verification System Prompt}
This prompt runs only when deterministic keyword matching proposes a structural news event.
\begin{Verbatim}[fontsize=\normalsize,breaklines=true,breakanywhere=true]
You verify whether a CANDIDATE news event (found by keyword match) is a genuine RELATIONSHIP-BREAKING event for a statistical-arbitrage PAIR. A relationship-
breaking event materially and durably changes
one leg's business so the pair's historical
co-movement no longer holds. Examples: a leg
being acquired / merged / taken private,
bankruptcy or going-concern doubt, delisting, a
trading halt on a leg, fraud / accounting
scandal / financial restatement, a severe
leg-specific regulatory action, or a
catastrophic earnings/guidance shock to ONE leg.

Routine mentions (the company merely commenting
on industry M&A, an article about a DIFFERENT
company's bankruptcy, ordinary earnings/analyst
chatter) are NOT critical.

Return ONLY JSON: {"event_type": "...", "affects_pair_relationship": true/false, "severity": "LOW/MEDIUM/HIGH/CRITICAL", "reasoning": "..."}
\end{Verbatim}

\subsection{Runtime Composition and User Payloads}
The preceding text is the fixed system component. Before each model call, dates are masked as step labels; price/news alignment retains the original calendar internally. Each training replay and test window starts at step 001. The shared runtime instruction is:
\begin{Verbatim}[fontsize=\normalsize,breaklines=true,breakanywhere=true]
Calendar dates are masked as step labels.
current_step identifies the current trading step
within this training or test window, starting at
step 001. Historical records use the same
numbering. step XXX masks a date outside the
decision window or without a mapped trading
step. Do not infer calendar dates. Use only the
supplied observations and settled past-step
feedback. Do not reconstruct historical market
outcomes from model pretraining or treat
predicted future outcomes as observed facts.
\end{Verbatim}
At runtime, pair selection appends a \texttt{COMPANY CONTEXT} block and, when nonempty, \texttt{PAIR QUALITY MEMORY}; daily direction appends the same blocks. Accepted training lessons are appended to the daily-direction prompt exactly as follows:
\begin{Verbatim}[fontsize=\normalsize,breaklines=true,breakanywhere=true]
COMPANY CONTEXT (business only):
- <ticker -- company name: business description>

PAIR QUALITY MEMORY (spread mean-reversion only):
- <PAIR>: <quality> (<note>)

Learned Decision Principles (transferable; from training reflection):
- <accepted transferable principle>
\end{Verbatim}
For relationship-break assessment, the engine appends \texttt{(no memory)} because it does not pass pair memory to that call. The application sends the observation as a JSON user message rather than interpolating it into the system prompt. The schemas below use \texttt{step XXX} as a template placeholder; runtime calls substitute the actual step number when available:
\begin{Verbatim}[fontsize=\normalsize,breaklines=true,breakanywhere=true]
Pair selection:
{"current_step": "step XXX",
 "top5_candidates": [<candidate summaries>]}

Daily direction:
{"current_step": "step XXX",
 "active_pair": ["A", "B"],
 "active_candidate": <spread, price, and news summary>,
 "lifecycle_state": "HEALTHY|WARNING|BREAK",
 "past_3_days_direction_memory":
   [{"step": "step XXX", <finalized counterfactual fields>}]}

Training reflection:
{"current_step": "step XXX",
 "decisions": [<action and all-action counterfactual PnL records>],
 "win_rate": <value>, "avg_pnl": <value>,
 "counterfactual_summary": <diagnostic counts>}

Relationship-break assessment:
{"current_step": "step XXX",
 "pair": ["A", "B"],
 "statistical_deterioration": [<factors>],
 "statistical_status": <status>,
 "recent_rolling_correlation": <value>,
 "half_life_days": <value>, "adf_pvalue": <value>,
 "cointegration_pvalue": <value>, "current_spread_z": <value>,
 "spread_volatility_ratio": <value>,
 "code_news_break_level": <status>,
 "past_week_news_summary": <summary>,
 "monitoring_event_type": <type>,
 "days_since_last_break_validation": <count>}

Critical-news verification:
{"current_step": "step XXX",
 "pair": ["A", "B"], "candidate_event": <event>,
 "past_week_news_summary": <summary>}
\end{Verbatim}

\section{Data Details}
\label{app:data}
The experiment windows and news come from TheFinAI's Herculean release on Hugging Face \citep{peng2026herculean}. The primary experiment uses ten U.S. equities with 27 training sessions (2 January--11 February 2025) and 75 test sessions (12 February--30 May 2025). Prices are obtained through yfinance and news through Finnhub; historical blocks are truncated causally, prior-session news enters only the current observation, and raw news is never persisted in memory.

A1 uses Herculean's 332-session panel of AAPL, ADBE, AMZN, GOOGL, META, MSFT, NVDA, and TSLA: 28 training sessions from 2 December 2024 to 13 January 2025, followed by 304 test sessions from 14 January 2025 to 31 March 2026, yielding 303 finalized transitions. A2 uses a pre-specified, non-overlapping 103-session window: 28 training sessions from 2 June to 11 July 2025 and 75 test sessions from 14 July to 27 October 2025. A3 and A4 retain the primary training and test date boundaries. Both A3 conditions share the single-round warmup and its reflection-memory initialization; only the full condition then enables the belief-update procedure. A4 retains policy-memory training but disables lifecycle-controlled reselection; the pair trajectory changes as a downstream result.

Curves are reconstructed as $E_t=\prod_{k\leq t}(1+r_k)$, with metrics reported at the precision shown in each table. The action at $t$ sees prior-session news and receives alternative next-session outcomes only after prices arrive; only the derived reflection is retained, using a $0.2\%$ deadband.

\section{Model List}
\label{app:models}
Table~\ref{tab:model-list} lists 11 model identifiers, yielding 13 evaluated configurations: seven API backbones, two Llama sizes, and two Qwen sizes with thinking and no-think. Llama 3.1 provides 8B and 70B instruction-tuned dense models; Qwen3 provides dense 4B and 32B models with a shared-weight thinking switch. This selection compares model scale and reasoning mode without treating architecture as an isolated causal factor.

\begin{table*}[!t]
\centering\fontsize{9}{11}\selectfont
\setlength{\tabcolsep}{4pt}
\begin{tabularx}{\textwidth}{@{}>{\raggedright\arraybackslash}p{0.39\textwidth}>{\raggedright\arraybackslash}Xl@{}}
\toprule
\textbf{Model} & \textbf{API alias or model identifier} & \textbf{Developer} \\
\midrule
GPT-4.1-mini \citep{openai2025gpt41} & \nolinkurl{gpt-4.1-mini} & OpenAI \\
GPT-5.4-mini \citep{openai2026gpt54small,openai2026gpt54card} & \nolinkurl{gpt-5.4-mini} & OpenAI \\
GPT-5.4-nano \citep{openai2026gpt54small,openai2026gpt54card} & \nolinkurl{gpt-5.4-nano} & OpenAI \\
o4-mini \citep{openai2025o4mini} & \nolinkurl{o4-mini} & OpenAI \\
DS \citep{deepseek2026v4} & \nolinkurl{deepseek-v4.1-flash} & DeepSeek \\
GF \citep{google2026gemini31flashlite} & \nolinkurl{gemini-3.1-flash-lite} & Google DeepMind \\
GL \citep{google2026gemini38flash} & \nolinkurl{gemini-3.8-flash} & Google DeepMind \\
\midrule
L8 \citep{meta2024llama31} & \nolinkurl{meta-llama/Llama-3.1-8B-Instruct} & Meta \\
L70 \citep{meta2024llama31} & \nolinkurl{meta-llama/Llama-3.1-70B-Instruct} & Meta \\
Q4N \citep{qwen2025qwen34b} & \nolinkurl{Qwen/Qwen3-4B} & Qwen Team \\
Q4T \citep{qwen2025qwen34b} & \nolinkurl{Qwen/Qwen3-4B} & Qwen Team \\
Q32N \citep{qwen2025qwen332b} & \nolinkurl{Qwen/Qwen3-32B} & Qwen Team \\
Q32T \citep{qwen2025qwen332b} & \nolinkurl{Qwen/Qwen3-32B} & Qwen Team \\
\bottomrule
\end{tabularx}
\caption{All 13 evaluated configurations and their 11 model identifiers, corresponding to Table~\ref{tab:main}. Thinking/no-think uses the same Qwen weights with different inference settings.}
\label{tab:model-list}
\end{table*}

\paragraph{Open-weight run settings.}
Qwen thinking is controlled by \texttt{enable\_thinking=true/false}; the server separates reasoning from final answer content, and only the final answer is parsed for actions. Llama uses its standard instruction-tuned interface. The recorded runs use server-default sampling, a 16,384-token output cap, a 600-second request timeout, and the run seed in model requests. Policy-memory training uses No-Beta sizing, with at most eight replay cycles; the trained policy is then frozen for testing. The reported L8 Beta run contains one recorded direction-decision fallback.

\section{Baseline Definitions}
\label{app:baselines}
The passive controls hold BMRN/META throughout the primary test and cross \ls, \sl, and Equal-Weight actions with both exposure settings. These controls instantiate fixed-direction pair trading \citep{gatev2006pairs,avellaneda2010statistical} and naive equal weighting \citep{demiguel2009naive}. 
For A1, both \method{} settings initially select AMZN/MSFT, and four controls hold that pair while crossing \ls{} and \sl{} with the corresponding exposure. For A2, four matched directional controls use the initial AAPL/MSFT pair under each exposure. A4 holds its initially selected BMRN/META pair throughout the 75-session test by disabling lifecycle-controlled replacement; the 23-session retry does not replace an active pair.

\textbf{Traditional rule-based: Engle--Granger + Z-score.}
This deterministic baseline uses the causally selected fixed pair and the Engle--Granger residual \citep{engle1987cointegration}. A 20-session rolling regression produces the hedge ratio and residual Z-score; the policy enters \sl{} at $z_t\geq1$, enters \ls{} at $z_t\leq-1$, and exits to \flat{} once $|z_t|\leq0.25$. It is evaluated under the same No-Beta and Beta exposure rules and receives no news, memory, or test-period fitting.

\textbf{RL: EG-DDQN.}
EG-DDQN retains the same fixed pair but replaces the threshold policy with a value-based learner. Its state contains eight causal numeric features---the residual Z-score and change, hedge ratio, residual volatility, two leg returns, five-session relative momentum, and return correlation---plus a one-hot current position; its actions are \ls{}, \flat{}, and \sl{}, and its reward is next-session portfolio return. The implementation follows a deep Q-network (DQN) \citep{mnih2015human} with Double-DQN action evaluation \citep{vanhasselt2016deep}, a 32-unit hidden layer, replay buffer, target network, and 600 training episodes over each experiment's fixed training trajectory. Networks are frozen throughout testing.

\textbf{LLM agent: \textsc{FinMem} (GPT-4.1-mini).}
We adapt \textsc{FinMem}'s layered-memory approach \citep{yu2025finmem} to fixed-pair trading by changing only its prompt, output, and evaluation layers. Test prompts use step labels and exclude next-step price changes; outcome-conditioned reflection is restricted to warmup.

\begin{table*}[t]
\centering\fontsize{9}{11}\selectfont
\setlength{\tabcolsep}{3pt}
\renewcommand{\arraystretch}{1.08}
\renewcommand{\tabularxcolumn}[1]{m{#1}}
\begin{tabularx}{\textwidth}{@{}>{\raggedright\arraybackslash}m{0.24\textwidth}*{7}{>{\centering\arraybackslash}X}@{}}
\toprule
& \multicolumn{2}{c}{\textbf{Available resources}} & \multicolumn{5}{c}{\textbf{Workflow}} \\
\cmidrule(lr){2-3}\cmidrule(lr){4-8}
\textbf{Method} & \textbf{Statistical\newline screening} & \textbf{News/\newline business\newline context} & \textbf{Training/\newline warmup} & \textbf{Raw news} & \textbf{Reflection} & \textbf{Lifecycle\newline switching} & \textbf{Global\newline policy\newline update} \\ 
\midrule 
Passive strategies & \yes & \no & \no & \no & \no & \no & \no \\
EG+Z & \yes & \no & \no & \no & \no & \no & \no \\
EG-DDQN & \yes & \no & \yes & \no & \no & \no & \no \\
\textsc{FinMem} adaptation & \yes & \yes & \yes & \yes & \yes & \no & \no \\
\method{}-A3 & \yes & \yes & \yes & \no & \yes & \yes & \no \\
\method{}-A4 & \yes & \yes & \yes & \no & \yes & \no & \yes \\
\method{} & \yes & \yes & \yes & \no & \yes & \yes & \yes \\
\bottomrule
\end{tabularx}
\caption{Resource and workflow comparison. Statistical screening and news/business context denote information available to each method; the remaining columns denote how it adapts or operates over time. \textsc{FinMem} retains raw news and uses reflection, whereas \method{} retains compact derived state rather than raw news. A3 disables the replay-gated global policy update after shared warmup; A4 disables lifecycle-controlled pair replacement.}
\label{tab:mechanism-comparison}
\end{table*}

\subsection{Adaptation Implementation Details}
\label{app:adaptation_details}
Policy consolidation proposes prompt edits from training reflection, evaluates each candidate by cumulative training return (the sum of daily returns), and accepts it only if this return improves; otherwise, the previous policy memory is retained. The update loop uses early stopping to bound consolidation. At test time, the learned global prompt is frozen, while daily reflection and lifecycle monitoring remain active.

A3 controls whether the replay-gated global policy update (the belief-update procedure) is enabled after a shared single-round warmup. Both conditions use the same training window, warmup protocol, reflection-memory initialization, backbone, exposure rule, and test protocol; the full condition then runs replay-gated prompt consolidation with rollback and early stopping, while the control retains warmup reflections without this procedure. Daily Counterfactual Reflection and lifecycle-controlled pair reselection remain active in both conditions.

Pair reselection is conditional on accumulated lifecycle evidence at the 23-session validation boundary. The boundary permits, but does not require, pair replacement each cycle; the pair trajectory is its downstream outcome.

\subsection{Token Usage}
\label{app:token_usage}
Table~\ref{tab:token_calls} reports approximate token ranges for the five call types in \method{}. These are per-call ranges; the calls occur at different frequencies and are not summed as a single daily step. Pair replacement is conditional on lifecycle evidence at each 23-step validation boundary.

\begin{table}[!htbp]
\centering\fontsize{9}{11}\selectfont
\setlength{\tabcolsep}{2pt}

\renewcommand{\tabularxcolumn}[1]{m{#1}}
\begin{tabularx}{\columnwidth}{@{}>{\centering\arraybackslash}X>{\centering\arraybackslash}m{0.22\columnwidth}>{\centering\arraybackslash}m{0.16\columnwidth}>{\centering\arraybackslash}m{0.22\columnwidth}@{}}
\toprule
\textbf{Call type} & \textbf{Input} & \textbf{Output} & \textbf{Total}\\
\midrule
Pair Selection & 3,000--5,000 & 150--300 & 3,200--5,300\\
Daily Action Decision & 2,500--3,500 & 150--250 & 2,700--3,750\\
Lifecycle Validation & 2,000--4,000 & 100--250 & 2,100--4,250\\
Critical-news Verification & 1,000--2,000 & 50--150 & 1,050--2,150\\
Counterfactual Reflection & 1,500--3,000 & 200--400 & 1,700--3,400\\
\bottomrule
\end{tabularx}
\caption{Approximate input, output, and total tokens per \method{} call. Ranges are reported independently.}
\label{tab:token_calls}
\end{table}

Table~\ref{tab:token_usage} reports aggregate chat-model usage for the GPT-4.1-mini A1 run. The test phase contains 303 finalized steps; including the preceding 28-session training phase, \method{} uses 1,762,999 tokens. Its test usage includes daily decisions, lifecycle validation, and conditional reselection. \textsc{FinMem} estimates are based on chat contexts and outputs and exclude embedding tokens.

\begin{table}[!htbp]
\centering\fontsize{9}{11}\selectfont
\setlength{\tabcolsep}{3pt}
\begin{tabular}{llr}
\toprule
\textbf{Scope} & \textbf{System} & \textbf{Total tokens}\\
\midrule
Each test step & \method{} & 3,785\\
 & \textsc{FinMem} & 15,124\\
\midrule
23-step cycle & \method{} & 87,054\\
 & \textsc{FinMem} & 350,168\\
\midrule
A1 full 303 steps & \method{} & 1,146,837\\
 & \textsc{FinMem} & 4,582,455\\

\bottomrule
\end{tabular}
\caption{Aggregate chat-model tokens with GPT-4.1-mini. The \method{} cycle estimate is its test total divided by 303 and multiplied by 23; \textsc{FinMem}'s cycle estimate averages the first 13 complete cycles. A1 training + full test includes training before the 303 finalized test steps.}
\label{tab:token_usage}
\end{table}

The 303-step test contains 13 complete 23-step cycles and four remaining steps. \textsc{FinMem}'s cycle average excludes those final four steps. Over the full test, \method{} uses approximately one quarter of \textsc{FinMem}'s chat tokens, including calls for lifecycle monitoring and conditional reselection.

\section{Detailed Model Architecture}
\label{app:architecture}

\begin{figure*}[!p]
\centering
\includegraphics[width=0.89\textwidth]{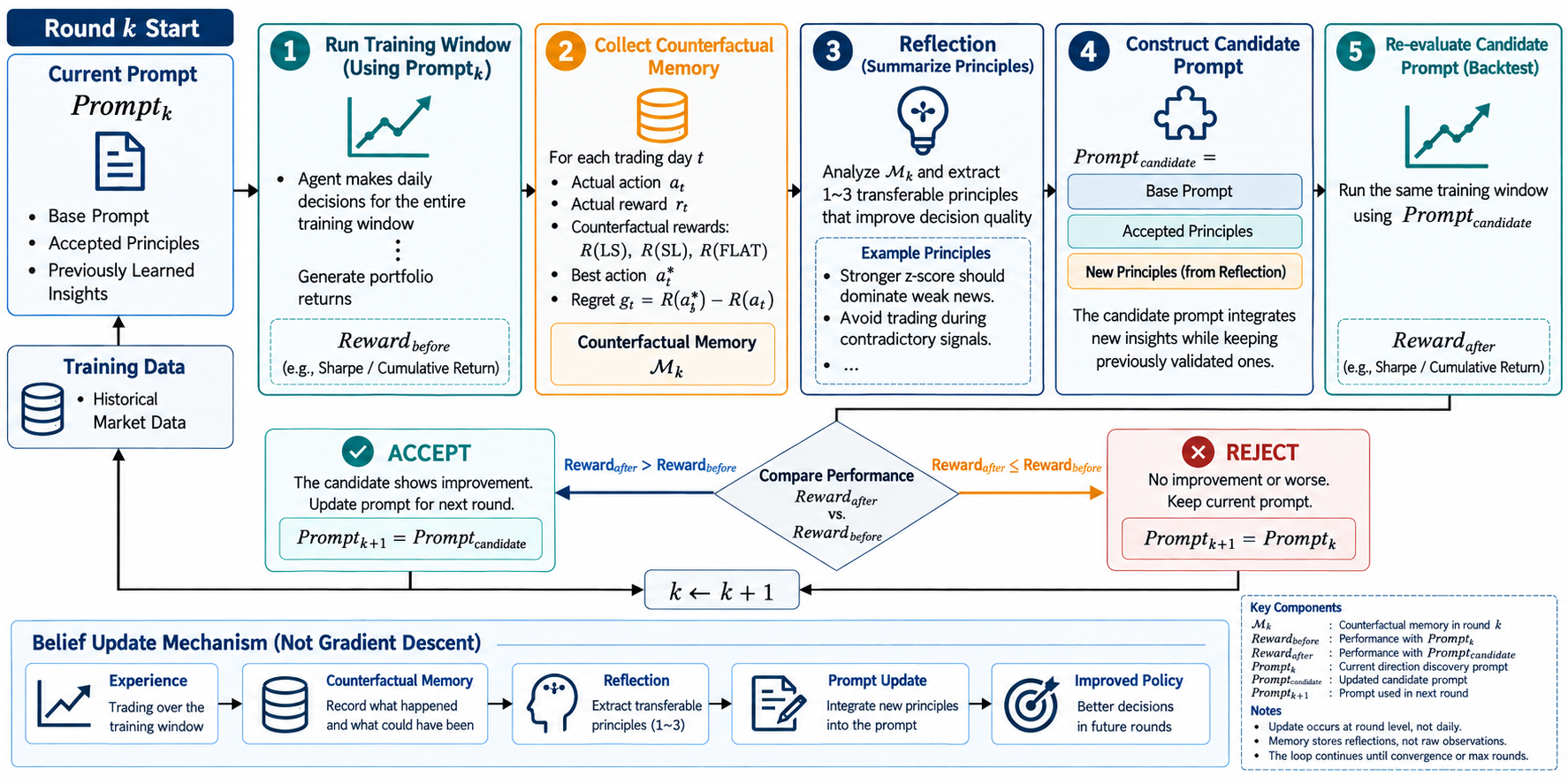}
\caption{(a) Policy-memory learning and validation.}
\label{fig:model-architecture-a}
\par\smallskip
\centering
\includegraphics[width=0.89\textwidth]{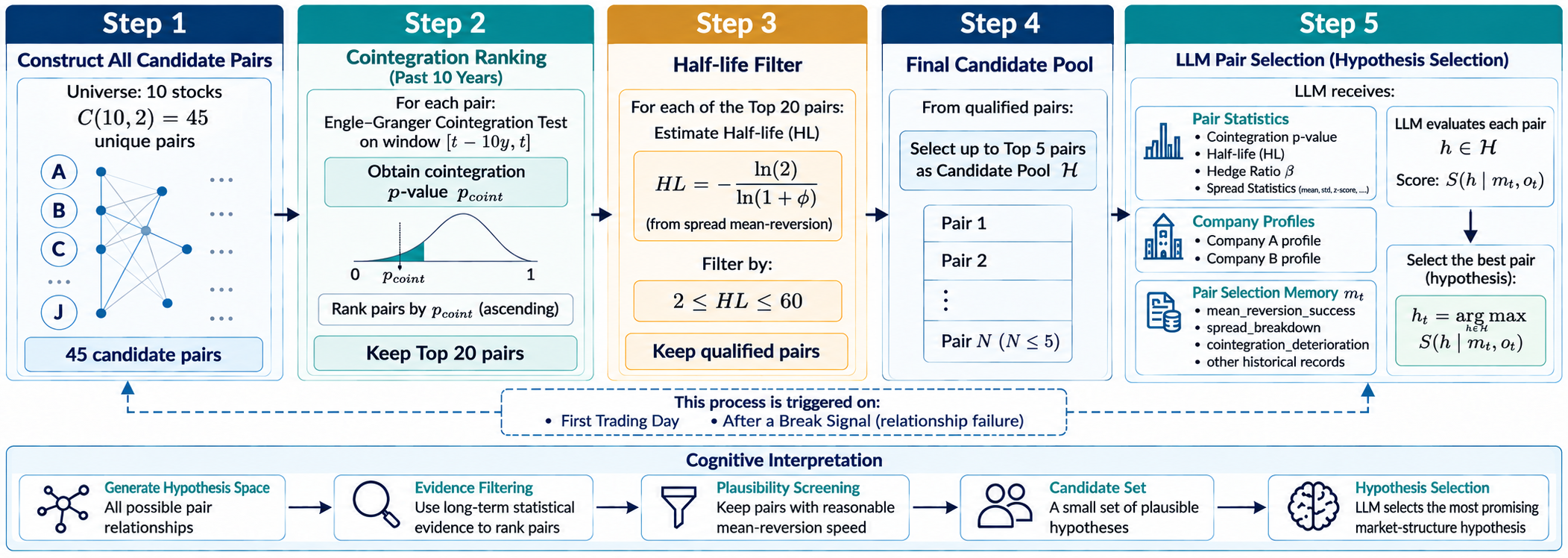}
\caption{(b) Pair selection as hypothesis selection.}
\label{fig:model-architecture-b}
\par\smallskip
\centering
\includegraphics[width=\textwidth]{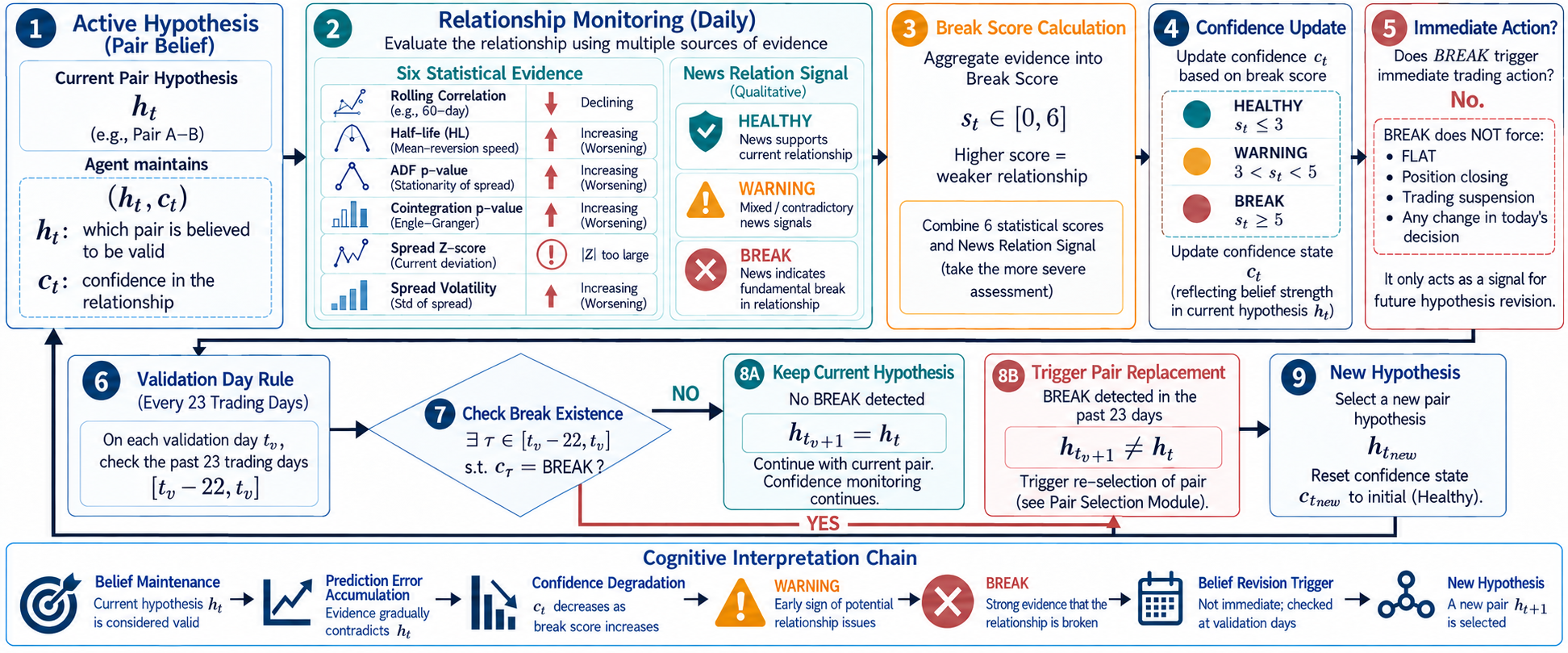}
\caption{(c) Lifecycle monitoring and belief revision.}
\label{fig:model-architecture-c}
\end{figure*}

This appendix provides a detailed view of the three components that implement \method{}'s adaptive workflow. Figure~\ref{fig:model-architecture-a} illustrates how training reflection and reward replay govern global policy updates; Figure~\ref{fig:model-architecture-b} details contextual pair selection using statistical screening together with news and company-business evidence; and Figure~\ref{fig:model-architecture-c} shows how lifecycle monitoring determines whether the active pair relationship should be retained or revised. Together, these diagrams complement the main workflow by making the information flow and update boundaries of each component explicit.



\section{Evaluation Metrics}
\label{app:metrics}
Let $r_t$ be the engine's one-session return for finalized transition $t$, $T$ the number of finalized transitions, and $E_0=1$, $E_t=\prod_{k=1}^{t}(1+r_k)$. Reported percentages are $100$ times their decimal values, and no transaction-cost estimate enters any measure.

\textbf{Trajectory metrics.}
Cumulative return is $C_T=E_T-1$. Annualized return is $(1+C_T)^{252/T}-1$, and annualized volatility is $\sqrt{252}\,s(r_{1:T})$, where $s$ is the sample standard deviation. With a zero per-session reference rate, annualized Sharpe ratio is
\begin{equation}
\mathrm{Sh}=\sqrt{252}\,\overline r/s(r_{1:T}).
\end{equation}
This follows the reward-to-variability definition of \citet{sharpe1994ratio}; \citet{lo2002statistics} motivates disclosing the annualization convention because temporal dependence can invalidate naive scaling. MDD is
\begin{equation}
\mathrm{MDD}=\min_{1\leq t\leq T}\left(E_t/\max_{0\leq u\leq t}E_u-1\right),
\end{equation}
following \citet{chekhlov2005drawdown}.

\textbf{Strict counterfactual action accuracy.}
For each finalized transition, let $a_t^{\star}=\arg\max_{a\in\mathcal{A}}r_t(a,h_t)$ and let $a_t$ be the final executed action. We compute
\[
\mathrm{Acc}_{\mathrm{CF}}=\frac{1}{T}\sum_{t=1}^{T}\mathbb{1}[a_t=a_t^{\star}].
\]
The denominator includes transitions with recorded realized returns. Ties are resolved in the fixed order Long--Short, Short--Long, Flat; this strict metric uses raw counterfactual returns without the reflection rubric's $0.2\%$ deadband.

\textbf{State diagnostics.}
Hypothesis switches are $N_{\mathrm{switch}}=\sum_{t=2}^{T}\mathbb{1}[h_t\neq h_{t-1}]$. Lifecycle replacements count validation boundaries at which an invalidated $h_t$ is retired. Training rounds, accepted updates, lifecycle occupancy, and action frequencies are direct event counts divided by their disclosed denominators when shown as rates. Counterfactual outcomes supervise reflection, and action accuracy is reported as a behavioral diagnostic.

\par\nopagebreak[4]\vspace{3.5ex}
\refstepcounter{section}
\label{app:financial-statistics}
\noindent{\large\bfseries \thesection\quad Financial and Statistical Calculations}\par\nopagebreak[4]\vspace{2.3ex}
Given $(o_t,x_t)$, the model returns \ls, \sl, or \flat. Statistical arbitrage denotes repeated long--short positions based on estimated relative-price structure; market neutrality offsets common exposure; cointegration denotes a stationary linear combination of non-stationary series; and mean reversion denotes its tendency to decay toward a long-run level \citep{engle1987cointegration,gatev2006pairs,avellaneda2010statistical}.

At selection date $t$, adjusted-close matrices are causally truncated to $\mathcal{W}_t=\{u:t-10\text{ years}\leq u\leq t\}$. Common positive-price sessions are retained, at least 60 observations are required, and all $\binom{10}{2}=45$ unordered pairs are enumerated. For each pair $(A,B)$, with $\ell_u^A=\log P_u^A$ and $\ell_u^B=\log P_u^B$, the first Engle--Granger stage estimates
\begin{equation}
\ell^A_u=\alpha+\beta^{\mathrm{sel}}\ell^B_u+\varepsilon_u,
\end{equation}
where $\widehat\beta^{\mathrm{sel}}$ defines the selection residual and the test null is no cointegration \citep{engle1987cointegration}. The residual is then evaluated with
\begin{equation}
\Delta\widehat\varepsilon_u=\gamma\widehat\varepsilon_{u-1}
 +\sum_{j=1}^{k}\delta_j\Delta\widehat\varepsilon_{u-j}+\nu_u,
\end{equation}
where the Akaike information criterion selects $k$ and the ADF null is $\gamma=0$ \citep{dickey1979unitroot,said1984unitroots}.

The 20 pairs with the lowest cointegration $p$-values undergo half-life screening. Mean-reversion speed is estimated from $\Delta\widehat\varepsilon_u=a+\phi\widehat\varepsilon_{u-1}+e_u$. For $-1<\widehat\phi<0$, the session half-life is
\begin{equation}
H=-\frac{\log 2}{\log(1+\widehat\phi)};
\end{equation}
otherwise $H=\infty$ \citep{uhlenbeck1930brownian,elliott2005pairs}. Among these 20 pairs, those with $2\leq H\leq60$ qualify; up to five qualified pairs enter the prompt in cointegration-$p$ order. Figure~\ref{fig:model-architecture-b} visualizes this division of labor.

\textbf{Exposure transformation.}
No-Beta sets the two share legs to equal magnitude. Beta instead estimates a separate rolling 20-session ordinary least squares (OLS) coefficient on log prices,
\begin{equation}
\widehat\beta^{\mathrm{risk}}_t=
\frac{\sum_{u\in L_t}(\ell_u^B-\bar \ell_t^B)(\ell_u^A-\bar \ell_t^A)}
{\sum_{u\in L_t}(\ell_u^B-\bar \ell_t^B)^2},
\end{equation}
where $L_t$ is the causal lookback. The effective magnitude is $\widetilde\beta_t=|\widehat\beta^{\mathrm{risk}}_t|$ when $.5\leq|\widehat\beta^{\mathrm{risk}}_t|\leq2$ and $1$ otherwise. Thus \ls{} uses share-sign ratio $(+1,-\widetilde\beta_t)$ and \sl{} reverses it before the engine scales gross exposure. For a given day, this position-sizing step changes realized $r_t(a,h_t)$ without overriding the action already selected.

\textbf{Lifecycle score.}
The statistical deterioration score is the sum $s_t=\sum_{j=1}^{6}s_{j,t}$ of the six fixed components in Table~\ref{tab:break-score}. Here $H_{\max}=60$, $p_{\max}=.05$, $z_t$ is the current spread z-score, and $v_t$ is current spread volatility divided by its reference value.
\begin{table}[!htbp]
\centering\fontsize{9}{11}\selectfont
\setlength{\tabcolsep}{3pt}
\begin{tabularx}{\columnwidth}{@{}lXr@{}}
\toprule
\textbf{Component} & \textbf{Condition $\rightarrow$ points} & \textbf{Max.} \\
\midrule
Correlation $\rho_t$ & $\rho_t<.35\!:\,4$; $[.35,.45)\!:\,3$; $[.45,.60)\!:\,2$; $[.60,.75)\!:\,1$; $\rho_t\geq.75\!:\,0$. & 4 \\
Half-life $H_t$ & Nonfinite or $H_t>240\!:\,3$; $(120,240]\!:\,2$; $(60,120]\!:\,1$; $H_t\leq60\!:\,0$. & 3 \\
ADF $p_t$ & $p_t>.20\!:\,3$; $(.10,.20]\!:\,2$; $(.05,.10]\!:\,1$; $p_t\leq.05\!:\,0$. & 3 \\
Cointegration $p_t$ & Same thresholds and points as ADF. & 3 \\
Spread $|z_t|$ & $|z_t|>3.5\!:\,2$; $(2.5,3.5]\!:\,1$; $|z_t|\leq2.5\!:\,0$. & 2 \\
Volatility ratio $v_t$ & $v_t>2\!:\,2$; $(1.5,2]\!:\,1$; $v_t\leq1.5\!:\,0$. & 2 \\
\midrule
Total $s_t$ & Sum of six component scores; no rescaling or clipping. & 17 \\
\bottomrule
\end{tabularx}
\caption{Implemented statistical break-score rubric, with $H_{\max}=60$ and both $p$-value cutoffs $.05$. The theoretical range is 0--17; the displayed experimental traces occupy 0--6. News is assessed separately.}
\label{tab:break-score}
\end{table}
The statistical label $q_t$ is \textsc{Healthy} for $s_t\leq3$, \textsc{Warning} for $3<s_t<5$, and \textsc{Break} for $s_t\geq5$. The final lifecycle state is the more severe of this label and an independent model-assessed news-relation state. The 20-session hedge lookback and 23-session validation boundary approximate one trading month; all score cutoffs follow the pre-registered code rubric and are constant across models, runs, and windows rather than tuned on test returns. At each 23-session boundary, Equation~\ref{eq:pair-reselection} uses the preceding cycle's break evidence. In the main experiment's initial selection on 12 February 2025, seven of 45 candidates qualify, giving the reported $7/45=15.6\%$ rate.

\FloatBarrier
\section{Behavioral Diagnostics and Case Study}
\label{app:case}
These diagnostics test whether the state factorization in Section~\ref{sec:method} makes distinct adaptation mechanisms observable. Table~\ref{tab:diagnostics} and Appendix~\ref{app:additional-results} separate hypothesis persistence, abstention behavior, lifecycle state, and the policy induced by accepted memory.

\begin{table}[!htbp]
\centering\fontsize{9}{11}\selectfont\setlength{\tabcolsep}{4pt}
\begin{tabular}{lrr|rr}
\toprule
&\multicolumn{2}{c}{\textbf{No-Beta}}&\multicolumn{2}{c}{\textbf{Beta}}\\
\textbf{Configuration}&\textbf{Switch}&\textbf{Flat}&\textbf{Switch}&\textbf{Flat}\\
\midrule
GPT-4.1-mini &2&7&2&8\\
GPT-5.4-mini &3&7&3&8\\
GPT-5.4-nano &2&11&2&15\\
o4-mini &2&15&1&23\\
DS &2&46&1&5\\
GF &2&22&3&15\\
GL &0&19&0&20\\
\midrule
L8 &1&0&1&4\\
L70 &1&26&1&25\\
Q4N &1&7&1&5\\
Q4T &1&1&1&1\\
Q32N &1&18&1&4\\
Q32T &1&10&1&2\\
\bottomrule
\end{tabular}
\caption{Actual held-pair switches and executed \flat{} actions in the recorded test trajectories. Initial selection is excluded; returning to a previous pair counts as a switch. Thinking/no-think follow Table~\ref{tab:main}.}
\label{tab:diagnostics}
\end{table}

\begin{figure*}[!p]

\begin{minipage}{\columnwidth}
\subsection{Drawdown Curves}
\end{minipage}
\par\smallskip

\centering
\includegraphics[width=\textwidth]{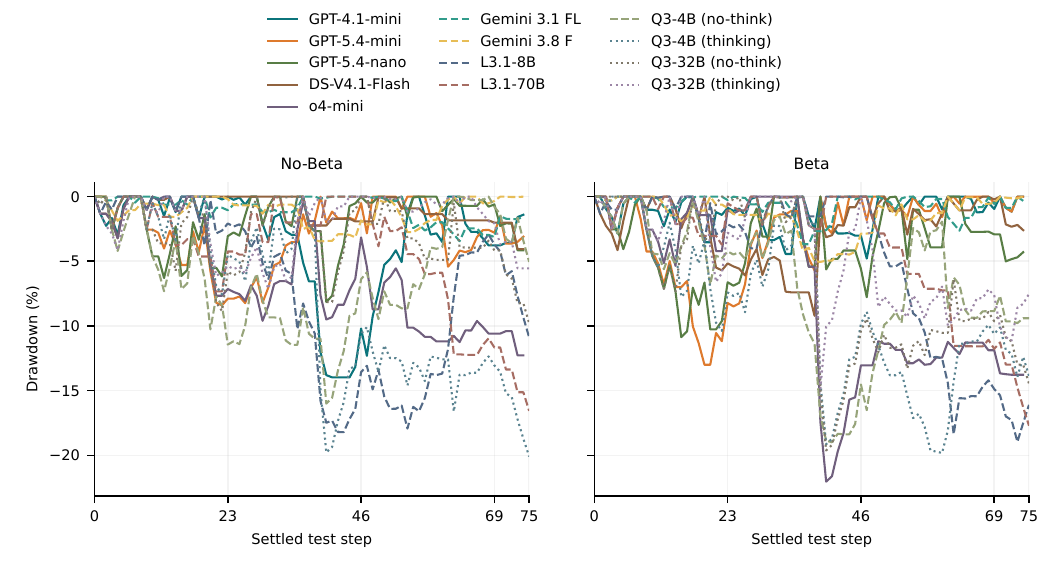}
\caption{Drawdown trajectories under No-Beta and Beta sizing.}
\label{fig:all-drawdown-all}

\par\smallskip
\begin{minipage}[t]{0.48\textwidth}
Beta's return gains seldom coincide with shallower drawdown: 11 of 13 configurations have deeper Beta MDD.
\end{minipage}\hfill
\begin{minipage}[t]{0.48\textwidth}
GPT-4.1-mini is the clearest exception, while o4-mini retains the deepest Beta trough (Figure~\ref{fig:all-drawdown-all}).
\end{minipage}

\par\medskip

\begin{minipage}{\columnwidth}
\subsection{Training Dynamics}
\end{minipage}
\par\smallskip

\centering
\includegraphics[width=\textwidth]{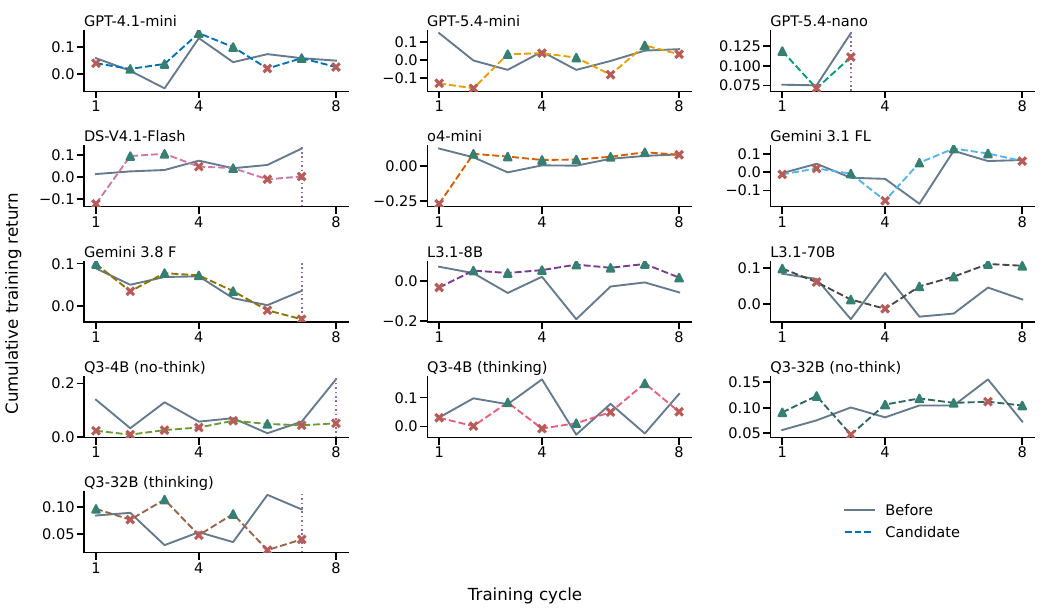}
\caption{Policy-training dynamics for the Beta evaluations. Triangles mark accepted updates, crosses rejected updates, and dotted lines early stopping. Llama/Qwen policy training uses No-Beta sizing.}
\label{fig:all-training-beta}

\par\smallskip
\begin{minipage}[t]{0.48\textwidth}
Training acceptance does not rank test performance: L8 accepts seven updates, yet Q32T Beta earns more with three.
\end{minipage}\hfill
\begin{minipage}[t]{0.48\textwidth}
Early stopping reflects validation of cumulative training return rather than guaranteed test success (Figure~\ref{fig:all-training-beta}).
\end{minipage}

\end{figure*}

\textbf{Validated memory measures selectivity rather than accumulation.}
L8 accepts seven of eight proposals, yet Q32T Beta earns more with three of seven. Likewise, o4-mini accepts six updates but loses under No-Beta, while GPT-5.4-nano profits with one (Figure~\ref{fig:all-training-beta}). Replay acceptance validates an edit locally, not its generalization.

\textbf{Policy adaptation and belief revision operate on different timescales.}
All 12 Llama/Qwen trajectories replace their pair once, at the second of three validation boundaries, despite Beta returns spanning $-6.3\%$ to $31.2\%$. Their transient \textsc{Break} states precede revision, separating daily evidence from pair replacement. Across all configurations, switches range from zero to three (Table~\ref{tab:diagnostics}; Figures~\ref{fig:all-pair-switch-all}--\ref{fig:pair-switch-beta}). Switch counts alone do not explain performance.

\textbf{Risk neutralization interacts with a learned policy.}
Beta improves both return and drawdown for GPT-4.1-mini, but L70's 18.7-point return gain accompanies deeper drawdown, as does o4-mini's recovery to positive return. Figures~\ref{fig:all-beta-effects-all}--\ref{fig:beta-effect-sharpe} show these exposure contrasts.

\textbf{Cases organized by error type.}
Each case identifies the model, exposure, experiment, and dates. One-step returns are decimal values; counterfactual alternatives condition on the same held pair. Subsequent correct actions show local recovery, not a causal effect of memory or proof that the pair remains valid.

\textbf{Type 1: Excessive abstention under conflicting signals.}
\setlength{\intextsep}{5pt}
\begin{casecard}{Case 1: Conflicting signals lead to a missed opportunity}
\textbf{Model:} GPT-5.4-nano\\
\textbf{Setting:} Main experiment, No-Beta test\\
\textbf{Date:} 8 April 2025 \hfill \textbf{Pair:} BMRN/META\\
\textbf{Evidence:} The agent chooses \flat{} (0.0000), while \sl{} would return 0.1267. Its rationale weighs a near-equilibrium long-horizon signal against a strongly negative trading-window signal, neutral news, and recent directional misses.\\
\textbf{Error analysis:} Conflicting evidence is resolved through abstention, missing the profitable direction ex post. The counterfactual supplies a concrete correction target; it does not establish that the opportunity was unambiguous before settlement.
\end{casecard}

\textbf{Type 2: Wrong-direction decisions and unreliable policy generalization.}
\begin{casecard}{Case 2: Continuation is favored before a reversal}
\textbf{Model:} GPT-5.4-nano\\
\textbf{Setting:} Main experiment, No-Beta test\\
\textbf{Dates:} 9--17 April 2025 \hfill \textbf{Pair:} BMRN/META\\
\textbf{Evidence:} On 9 April, the agent favors short-term divergence and relative weakness over mean reversion, choosing \sl{} ($-0.0564$) instead of \ls{} (0.0564). It then chooses the optimal \ls{} action for six sessions: 0.0053 on 10 April, 0.0249 on 11 April, and a four-session sum of 0.0796 on 14--17 April.\\
\textbf{Error analysis:} The losing decision extrapolates continuation into a reversal. Later rationales refer to recent misses and support mean reversion, consistent with memory-conditioned correction. The profitable sequence demonstrates local recovery without isolating memory's causal contribution.
\end{casecard}

\begin{casecard}{Case 3: A direction change repeats the error}
\textbf{Model:} Q4T\\
\textbf{Setting:} Main experiment, No-Beta test\\
\textbf{Dates:} 8--17 April 2025 \hfill \textbf{Pair:} BMRN/META\\
\textbf{Evidence:} \ls{} loses 0.1267 on 8 April; the switch to \sl{} loses 0.0563 on 9 April. Each rationale invokes recent counterfactual successes to support its direction, but the next outcome favors the opposite action. The six subsequent \ls{} actions are optimal; the 11 April rationale explicitly cites the correct 10 April decision.\\
\textbf{Error analysis:} Recent success is generalized into directional support despite changing momentum and reversion signals. Flipping direction therefore does not itself correct the decision rule. The later recovery coexists with a $-4.13\%$ full-test return, separating local adaptation from overall success.
\end{casecard}

\begin{casecard}{Case 4: Plausible policy guidance fails replay}
\textbf{Model:} Q32T\\
\textbf{Setting:} Main experiment, shared No-Beta training for the Beta evaluation\\
\textbf{Dates:} 2 January--11 February 2025 (replayed in cycles 2--3)\\
\textbf{Evidence:} Cycle 2 proposes momentum-following and selective-abstention principles, but cumulative training return falls from 0.0891 to 0.0763, so the edit is rejected. Cycle 3's proposed guidance improves cumulative training return from 0.0288 to 0.1139 and is accepted. Three of seven proposals are retained before early stopping.\\
\textbf{Error analysis:} A plausible verbal reflection can produce worse action guidance. Replay blocks this harmful proposal rather than treating reflection as improvement; acceptance remains evidence of in-window benefit, not guaranteed transfer.
\end{casecard}

A3 provides the corresponding controlled comparison: GPT-5.4-mini, No-Beta, over 12 February--30 May 2025 earns $2.7\%$ without post-warmup policy updates versus $24.3\%$ with them (Table~\ref{tab:a3}). Retaining reflection alone does not recover the benefit of validated policy consolidation; this aggregate contrast does not identify the cause of any single decision error.

\textbf{Type 3: Action--hypothesis attribution ambiguity.}
\begin{casecard}{Case 5: Successful actions do not validate a pair}
\textbf{Model:} Q32T\\
\textbf{Setting:} Main experiment, Beta test\\
\textbf{Dates:} 8--22 April 2025\\
\textbf{Pairs:} BMRN/META, then BMRN/CRM on 21 April\\
\textbf{Evidence:} BMRN/META records \textsc{Break} on 8--10 April, although its 10 April \ls{} action is profitable (0.0053). At the 21 April validation, three earlier break records trigger replacement despite a current score of 3 (\textsc{Healthy}). The new pair's first two \ls{} decisions lose 0.0182 and 0.0234; the 22 April rationale still cites profitable 16--17 April decisions on the previous pair.\\
\textbf{Error analysis:} Action success on one pair is insufficient evidence for its relationship validity or for the same direction on a replacement pair. The switch follows the accumulated-evidence rule, while the subsequent losses are wrong-direction decisions conditional on the new pair. Neither outcome alone establishes that revision was unnecessary or that the new relationship was invalid.
\end{casecard}

A4 isolates the revision mechanism: GPT-4.1-mini, Beta, over 12 February--30 May 2025 earns $2.8\%$ with lifecycle-controlled reselection disabled versus $30.6\%$ with it (Table~\ref{tab:a4}). Daily action correction alone does not replace relationship maintenance; the controlled contrast supports their distinct roles without attributing every post-switch gain or loss to pair selection.

\captionsetup{hypcap=false}
\FloatBarrier
\raggedbottom
\makeatletter
\setlength{\@dblfptop}{0pt}
\setlength{\@dblfpsep}{12pt}
\makeatother
\section{Additional Visual Results}
\label{app:additional-results}
The plots compare all 13 configurations in Table~\ref{tab:main}, including Llama/Qwen. Single-column diagnostics precede the wider multi-panel comparisons.

\noindent\begin{minipage}{\columnwidth}
\subsection{Effect of Beta Hedging}

\centering
\includegraphics[width=\columnwidth]{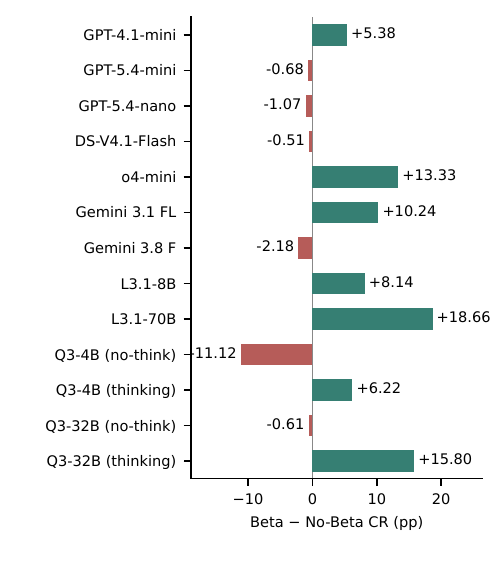}
\captionof{figure}{Beta minus No-Beta cumulative return.}
\label{fig:all-beta-effects-all}

\end{minipage}
\par\medskip

\noindent\begin{minipage}{\columnwidth}

\centering
\includegraphics[width=\columnwidth]{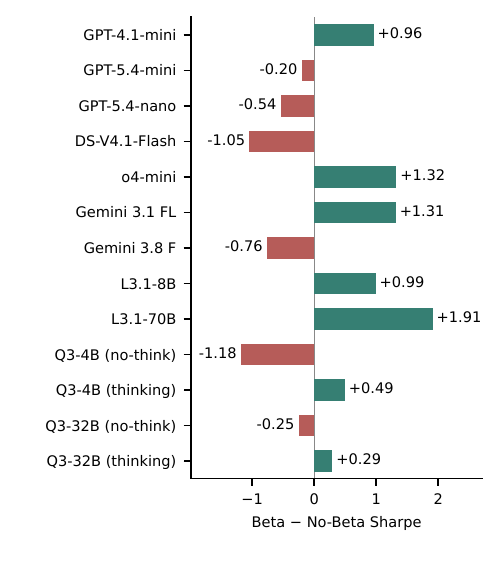}
\captionof{figure}{Beta minus No-Beta Sharpe ratio.}
\label{fig:beta-effect-sharpe}

\par\smallskip
Beta has higher return and Sharpe for seven configurations and lower values for six. Both Llama runs favor Beta, whereas both non-thinking Qwen configurations favor No-Beta, reinforcing exposure-dependent performance (Figures~\ref{fig:all-beta-effects-all}--\ref{fig:beta-effect-sharpe}).
\end{minipage}
\par\medskip

\noindent\begin{minipage}{\columnwidth}
\subsection{Pair-Selection Heatmaps}

\centering
\includegraphics[width=\columnwidth]{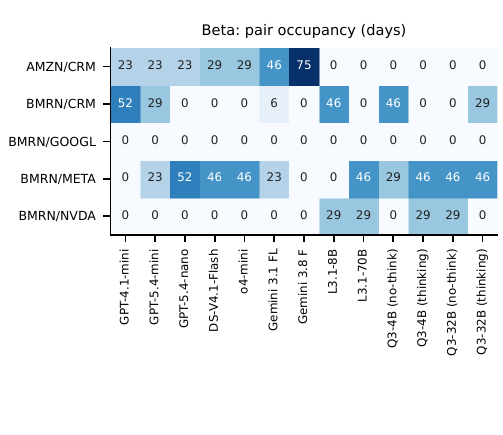}
\captionof{figure}{Pair-selection occupancy under Beta sizing.}
\label{fig:all-pair-heatmap-beta}

\end{minipage}
\par\medskip

\noindent\begin{minipage}{\columnwidth}

\centering
\includegraphics[width=\columnwidth]{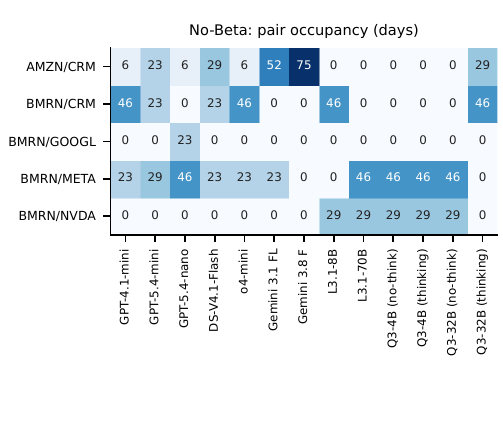}
\captionof{figure}{Pair-selection occupancy under No-Beta sizing.}
\label{fig:all-pair-heatmap-nobeta}

\par\smallskip
The open-weight runs concentrate on BMRN-anchored pairs, whereas Gemini mainly holds AMZN/CRM. Shared screening thus permits distinct pair trajectories across backbones (Figures~\ref{fig:all-pair-heatmap-beta}--\ref{fig:all-pair-heatmap-nobeta}).
\end{minipage}
\par\medskip

\noindent\begin{minipage}{\columnwidth}
\subsection{Action Distributions}

\centering
\includegraphics[width=\columnwidth]{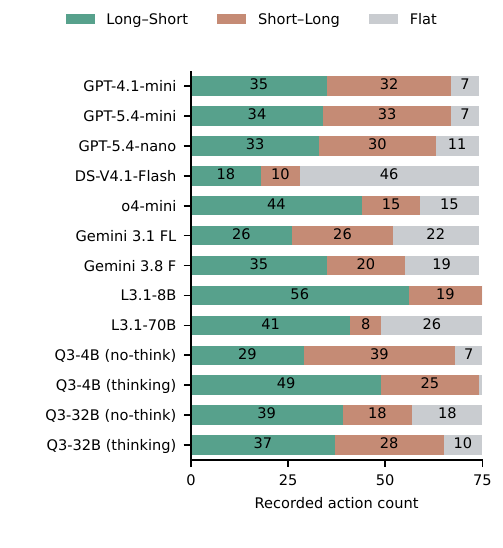}
\captionof{figure}{Action distribution under No-Beta sizing.}
\label{fig:all-actions-all}

\end{minipage}
\par\medskip

\noindent\begin{minipage}{\columnwidth}

\centering
\includegraphics[width=\columnwidth]{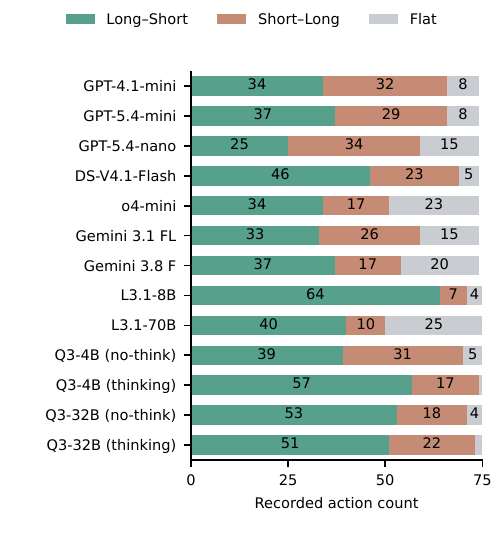}
\captionof{figure}{Action distribution under Beta sizing.}
\label{fig:actions-beta}

\par\smallskip
The Llama Beta runs contrast directional commitment with abstention: 8B executes \ls{} 64 times, whereas 70B records 25 \flat{} actions yet earns more. More directional activity does not imply better performance (Figures~\ref{fig:all-actions-all}--\ref{fig:actions-beta}).
\end{minipage}
\par\medskip

\noindent\begin{minipage}{\columnwidth}
\subsection{Pair-Switch Comparison}

\centering
\includegraphics[width=\columnwidth]{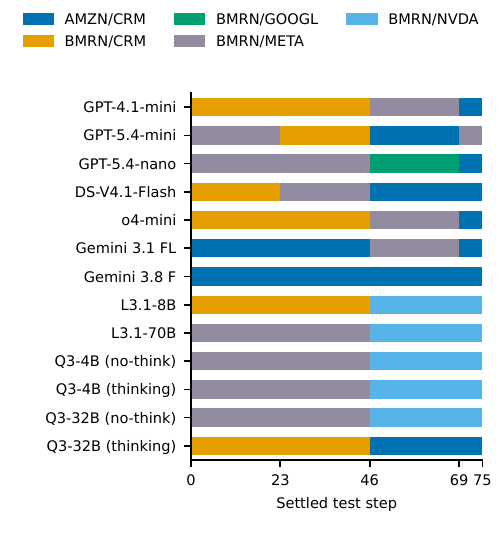}
\captionof{figure}{Pair trajectory under No-Beta sizing.}
\label{fig:all-pair-switch-all}

\end{minipage}
\par\medskip

\noindent\begin{minipage}{\columnwidth}

\centering
\includegraphics[width=\columnwidth]{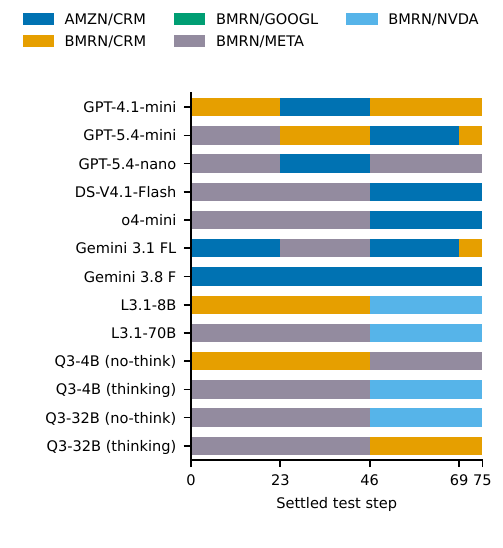}
\captionof{figure}{Pair trajectory under Beta sizing.}
\label{fig:pair-switch-beta}

\par\smallskip
Boundaries indicate actual held-pair changes. All Llama/Qwen runs switch once despite divergent returns. GPT-5.4-mini changes pair three times under each exposure without leading returns, while GL never switches (Figures~\ref{fig:all-pair-switch-all}--\ref{fig:pair-switch-beta}).
\end{minipage}
\par\medskip

\begin{figure*}[!tp]
\begin{minipage}[t]{0.48\textwidth}
\subsection{Counterfactual Action Accuracy}

\centering
\includegraphics[width=0.92\linewidth]{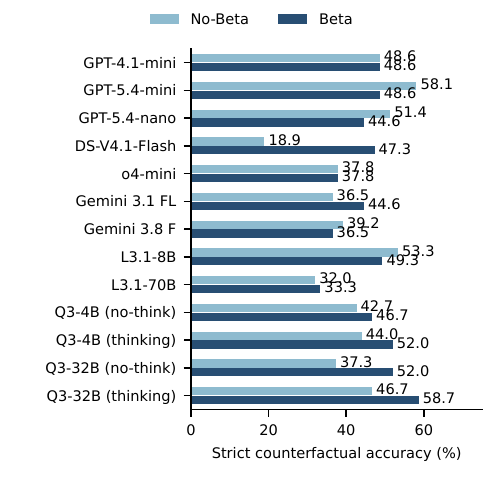}
\captionof{figure}{Strict counterfactual action accuracy under No-Beta and Beta sizing.}
\label{fig:all-accuracy-all}

\end{minipage}\hfill
\begin{minipage}[t]{0.48\textwidth}
\vspace{2.5ex}
Strict counterfactual action accuracy is defined in Appendix~\ref{app:metrics}. The leaders differ by exposure: GPT-5.4-mini under No-Beta (58.1\%) and Q32T under Beta (58.7\%). DS's low No-Beta accuracy coexists with positive return: accuracy ignores payoff magnitude, and \flat{} is rarely the strict argmax (Figure~\ref{fig:all-accuracy-all}).
\end{minipage}
\par\medskip
\begin{minipage}{\columnwidth}
\subsection{Return and Sharpe Comparison}
\end{minipage}
\par\smallskip

\centering
\includegraphics[width=0.96\textwidth]{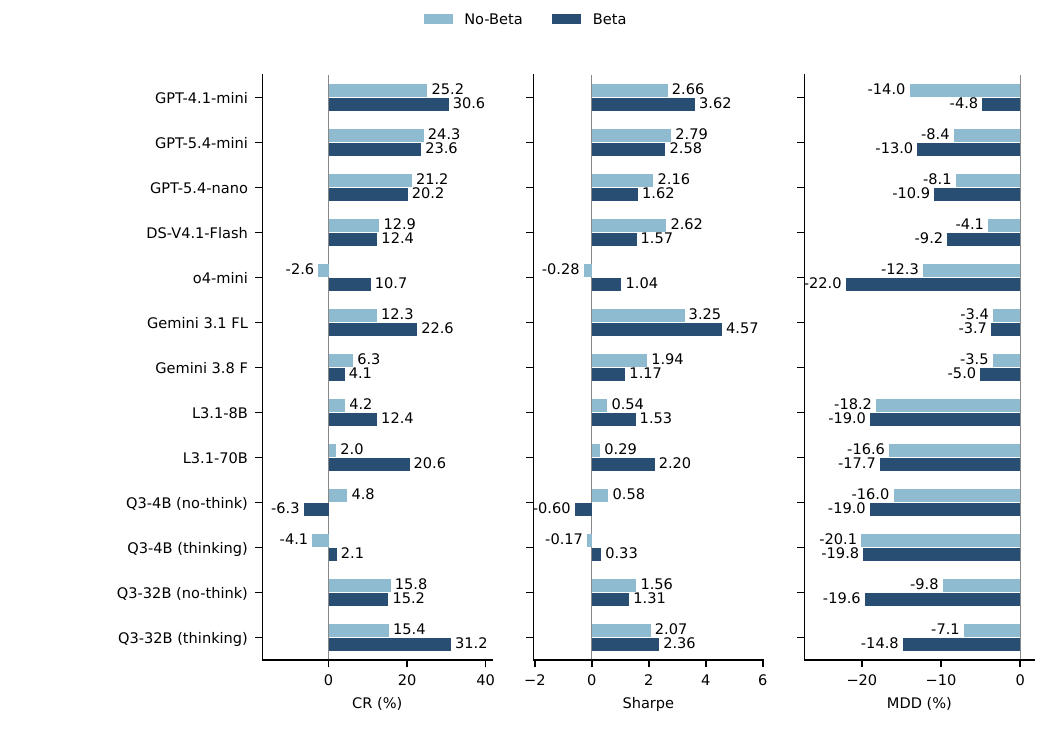}
\caption{Cumulative return (CR), Sharpe ratio, and maximum drawdown (MDD) under No-Beta and Beta sizing.}
\label{fig:all-metrics-all}

\par\smallskip
\begin{minipage}[t]{0.48\textwidth}
Return leadership changes with exposure: GPT-4.1-mini leads No-Beta, while Q32T leads Beta.
\end{minipage}\hfill
\begin{minipage}[t]{0.48\textwidth}
Three negative conditions remain across o4-mini and the Q4 modes, so gains are broad but not universal. GF leads Sharpe under both exposures despite lower returns than the return leaders; DS also combines moderate No-Beta return with comparatively strong risk-adjusted performance (Figure~\ref{fig:all-metrics-all}).
\end{minipage}
\end{figure*}

\begin{figure*}[!p]

\begin{minipage}{\columnwidth}
\subsection{Break-Score Evolution}
\end{minipage}
\par\smallskip

\centering
\includegraphics[width=\textwidth]{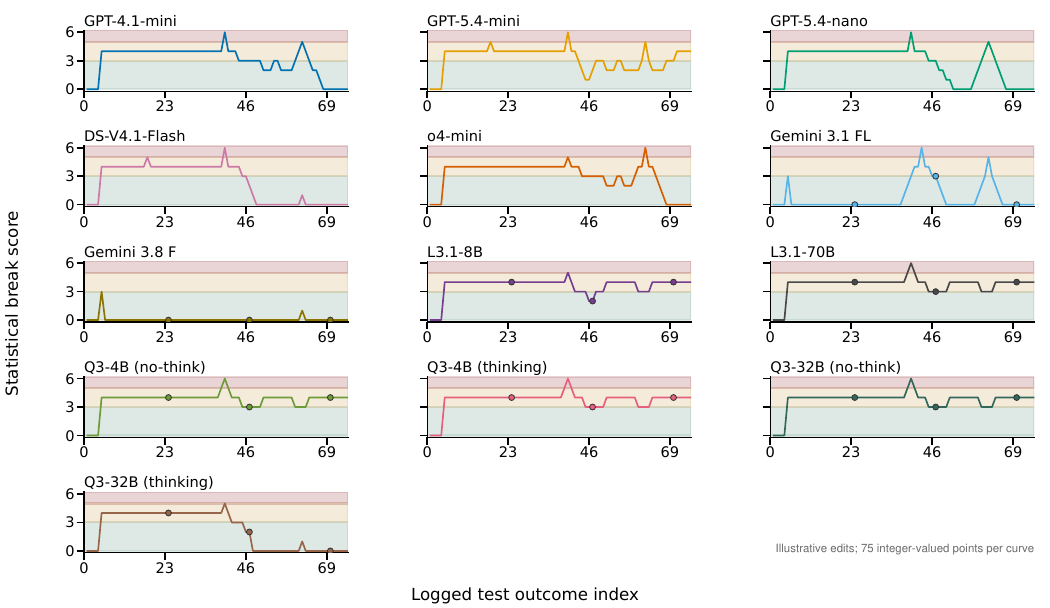}
\caption{Break scores under No-Beta sizing; dots mark validation days and shaded bands show lifecycle thresholds.}
\label{fig:all-break-scores-nobeta}

\par\medskip

\centering
\includegraphics[width=\textwidth]{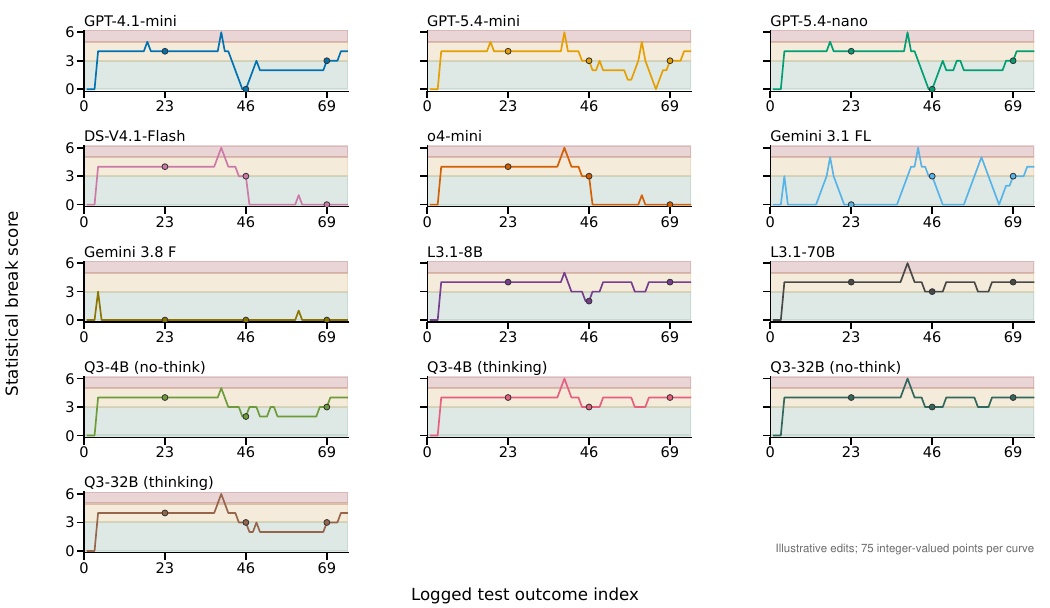}
\caption{Break scores under Beta sizing; markers and bands follow Figure~\ref{fig:all-break-scores-nobeta}.}
\label{fig:all-break-scores-beta}

\par\smallskip
\begin{minipage}[t]{0.48\textwidth}
Transient \textsc{Break} states occur across backbones, while GL remains \textsc{Healthy}.
\end{minipage}\hfill
\begin{minipage}[t]{0.48\textwidth}
In the open-weight runs, accumulated break evidence triggers one validation-boundary replacement; daily score recovery does not erase earlier evidence (Figures~\ref{fig:all-break-scores-nobeta}--\ref{fig:all-break-scores-beta}; Case 5).
\end{minipage}

\end{figure*}

\end{document}